\documentclass[10pt,twocolumn,letterpaper]{article}

\usepackage[pagenumbers]{cvpr} 

\usepackage{graphicx}
\usepackage{amsmath}
\usepackage{amssymb}
\usepackage{booktabs}
\usepackage{multirow}
\usepackage{array}
\usepackage{microtype}
\usepackage{enumitem}
\newcolumntype{H}{>{\setbox0=\hbox\bgroup}c<{\egroup}@{}}

\usepackage[pagebackref,breaklinks,colorlinks]{hyperref}

\usepackage[capitalize]{cleveref}
\crefname{section}{Sec.}{Secs.}
\Crefname{section}{Section}{Sections}
\Crefname{table}{Table}{Tables}
\crefname{table}{Tab.}{Tabs.}

\def\cvprPaperID{12901} 
\def\confName{CVPR}
\def\confYear{2023}

\newcommand{\ours}[0]{VectorFusion}
\newcommand{\ourtitle}[0]{\ours{}: Text-to-SVG by Abstracting Pixel-Based Diffusion Models}

\newcommand{\x}[0]{\mathbf{x}}
\newcommand{\z}[0]{\mathbf{z}}

\newcommand\blfootnote[1]{%
  \begingroup
  \renewcommand\thefootnote{}\footnote{#1}%
  \addtocounter{footnote}{-1}%
  \endgroup
}

\newcommand{\tightcaption}[1]{\vspace{-6mm} \caption{#1} \vspace{1mm}}

\begin{document}

\title{\ourtitle{}}

\author{Ajay Jain$^*$ \qquad Amber Xie$^*$ \qquad Pieter Abbeel \\
UC Berkeley \quad {\tt\small {\{ajayj,amberxie,pabbeel\}@berkeley.edu}}
}

\twocolumn[{
\vspace{-1cm}
\maketitle
\vspace{-1cm}
\begin{center}
    \includegraphics[width=\linewidth,trim={4mm 6mm 4mm 4mm},clip]{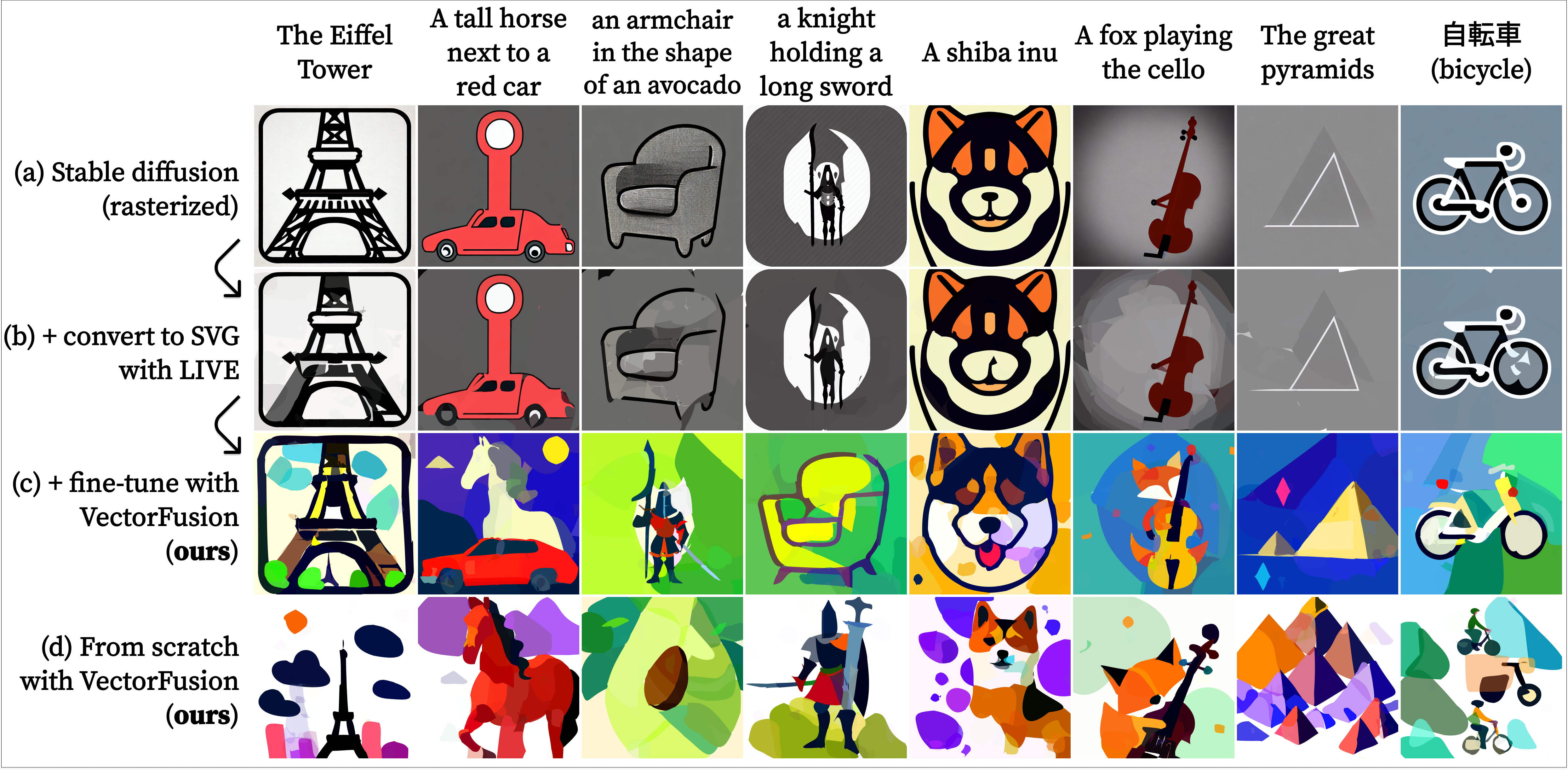}
    \captionof{figure}{Text-to-SVG with \ours{}. When (a) raster graphics sampled from Stable Diffusion are (b) auto-traced, they lose details that are hard to represent within the constraints of the abstraction. (c-d) \ours{} improves fidelity and consistency with the caption by directly optimizing paths with a distillation-based diffusion loss. Find videos and more results at~\href{https://ajayj.com/vectorfusion}{https://ajayj.com/vectorfusion}.}
    \label{fig:sd_comparison}
\end{center}
}]

\begin{abstract}
    \vspace{-0.25cm}
   Diffusion models have shown impressive results in text-to-image synthesis. Using massive datasets of captioned images, diffusion models learn to generate raster images of highly diverse objects and scenes. However, designers frequently use vector representations of images like Scalable Vector Graphics (SVGs) for digital icons or art. Vector graphics can be scaled to any size, and are compact. We show that a text-conditioned diffusion model trained on pixel representations of images can be used to generate SVG-exportable vector graphics. We do so without access to large datasets of captioned SVGs. By optimizing a differentiable vector graphics rasterizer, our method, \ours{}, distills abstract semantic knowledge out of a pretrained diffusion model. Inspired by recent text-to-3D work, we learn an SVG consistent with a caption using Score Distillation Sampling. To accelerate generation and improve fidelity, \ours{} also initializes from an image sample. Experiments show greater quality than prior work, and demonstrate a range of styles including pixel art and sketches.\blfootnote{$^*$Equal contribution}
\end{abstract}

\vspace{-1cm}
\section{Introduction}
\label{sec:intro}

Graphic designers and artists often express concepts in an abstract manner, such as composing a few shapes and lines into a pattern that evokes the essence of a scene. Scalable Vector Graphics (SVGs) provide a declarative format for expressing visual concepts as a collection of primitives. Primitives include B\'ezier curves, polygons, circles, lines and background colors. SVGs are the defacto format for exporting graphic designs since they can be rendered at arbitrarily high resolution on user devices, yet are stored and transmitted with a compact size, often only tens of kilobytes. Still, designing vector graphics is difficult, requiring knowledge of professional design tools.

\newcommand{\teaserwidthb}{.22\textwidth}
\newcommand{\teaserwidthd}{.19\textwidth}
\newcommand{\teaserwidthc}{.16\textwidth}
\newcommand{\betweencols}{\,\,\,}

\renewcommand{\tightcaption}[1]{\vspace{-5mm} \caption{#1} \vspace{1mm}}
\begin{figure*}
 \vspace{-0.5cm}
\captionsetup[subfigure]{labelformat=empty}
\centering
\resizebox{0.95\textwidth}{!}{%
\begin{tabular}{@{}c@{\quad}c@{\quad}c@{\quad}c@{}}
    \begin{subfigure}[t]{\teaserwidthb}
    \includegraphics[width=\columnwidth]{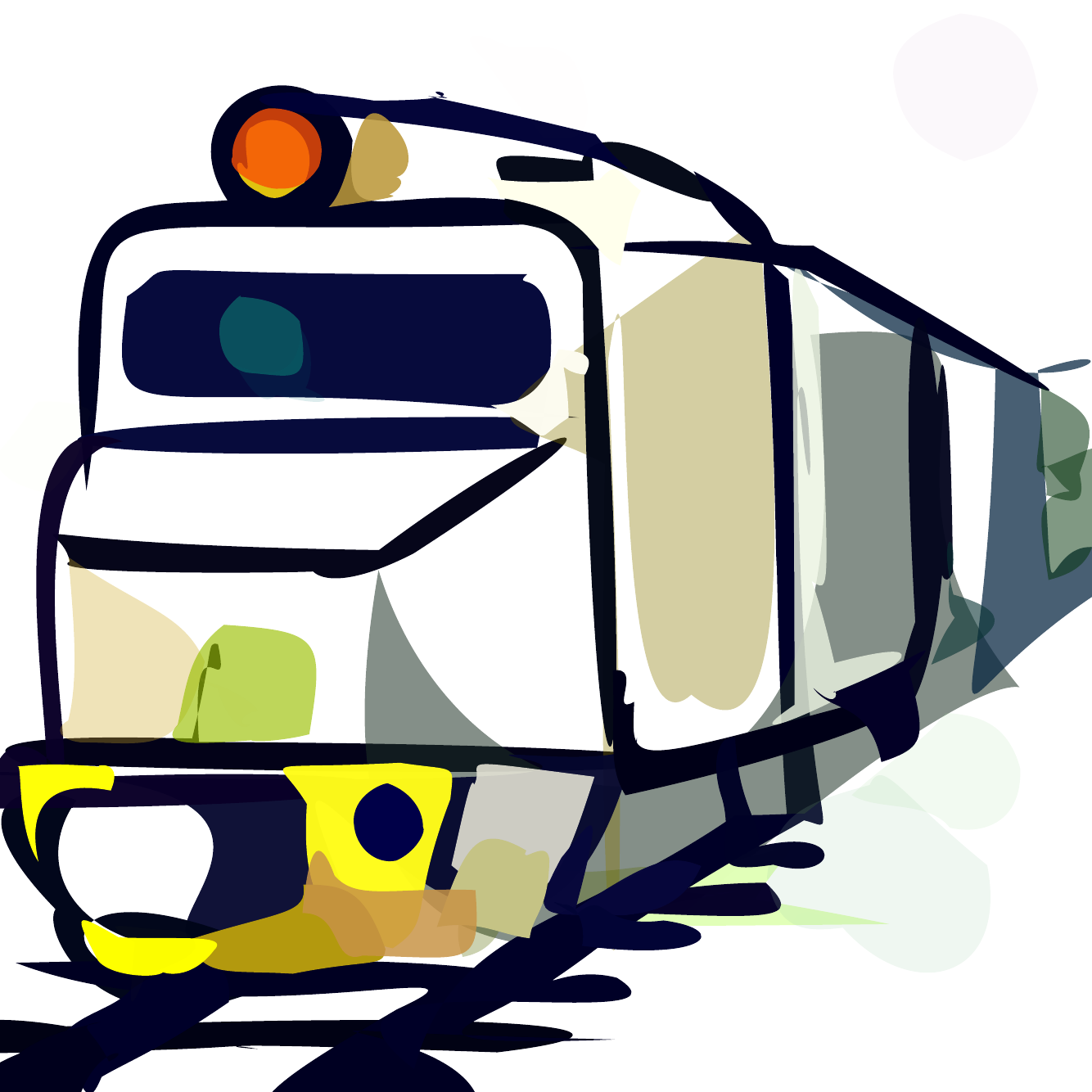}
    \tightcaption{\scriptsize{a train*}}
    \end{subfigure} &
    \begin{subfigure}[t]{\teaserwidthb}
    \includegraphics[width=\columnwidth]{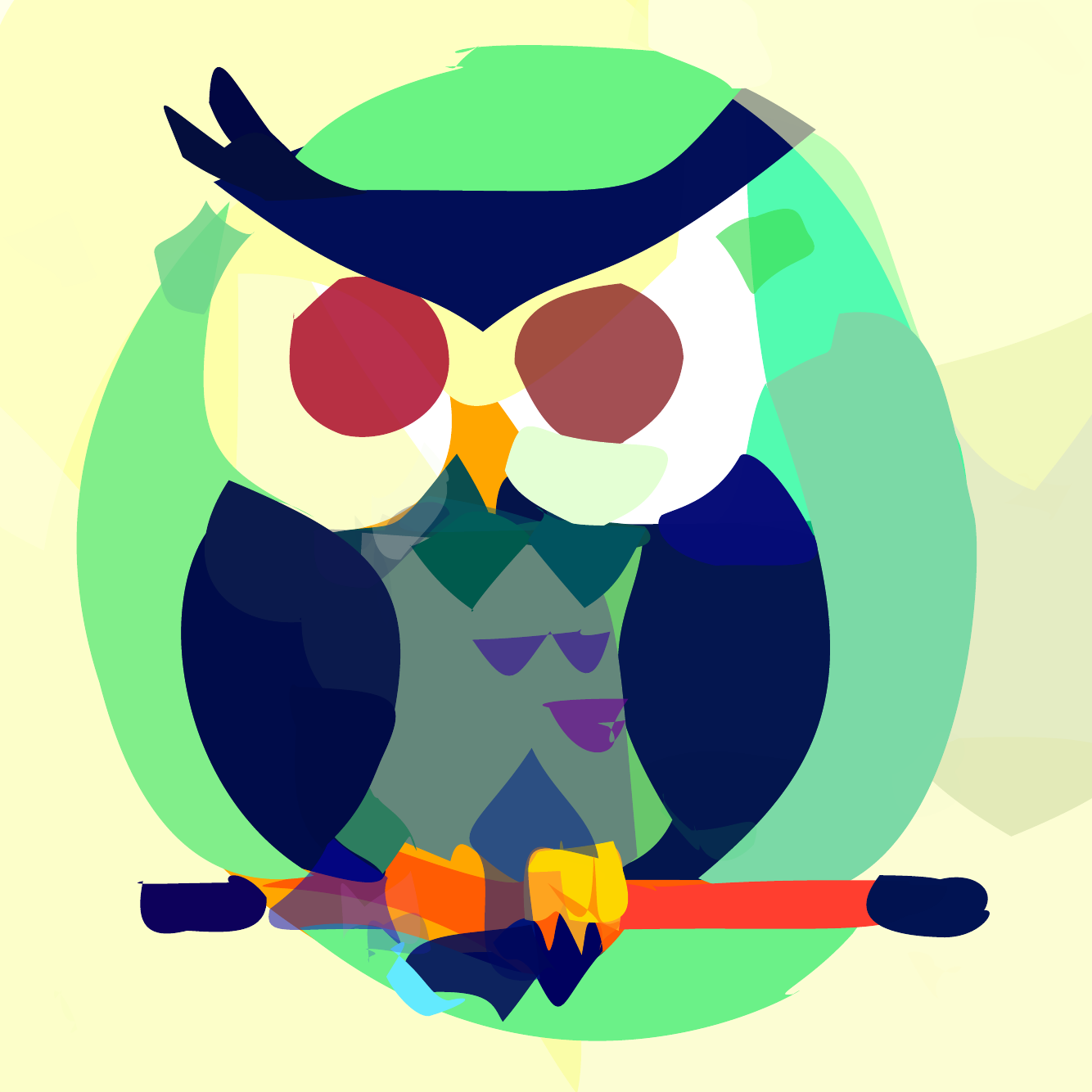}
    \tightcaption{\scriptsize{an owl standing on a wire*}}
    \end{subfigure} &
    \begin{subfigure}[t]{\teaserwidthb}
    \includegraphics[width=\columnwidth]{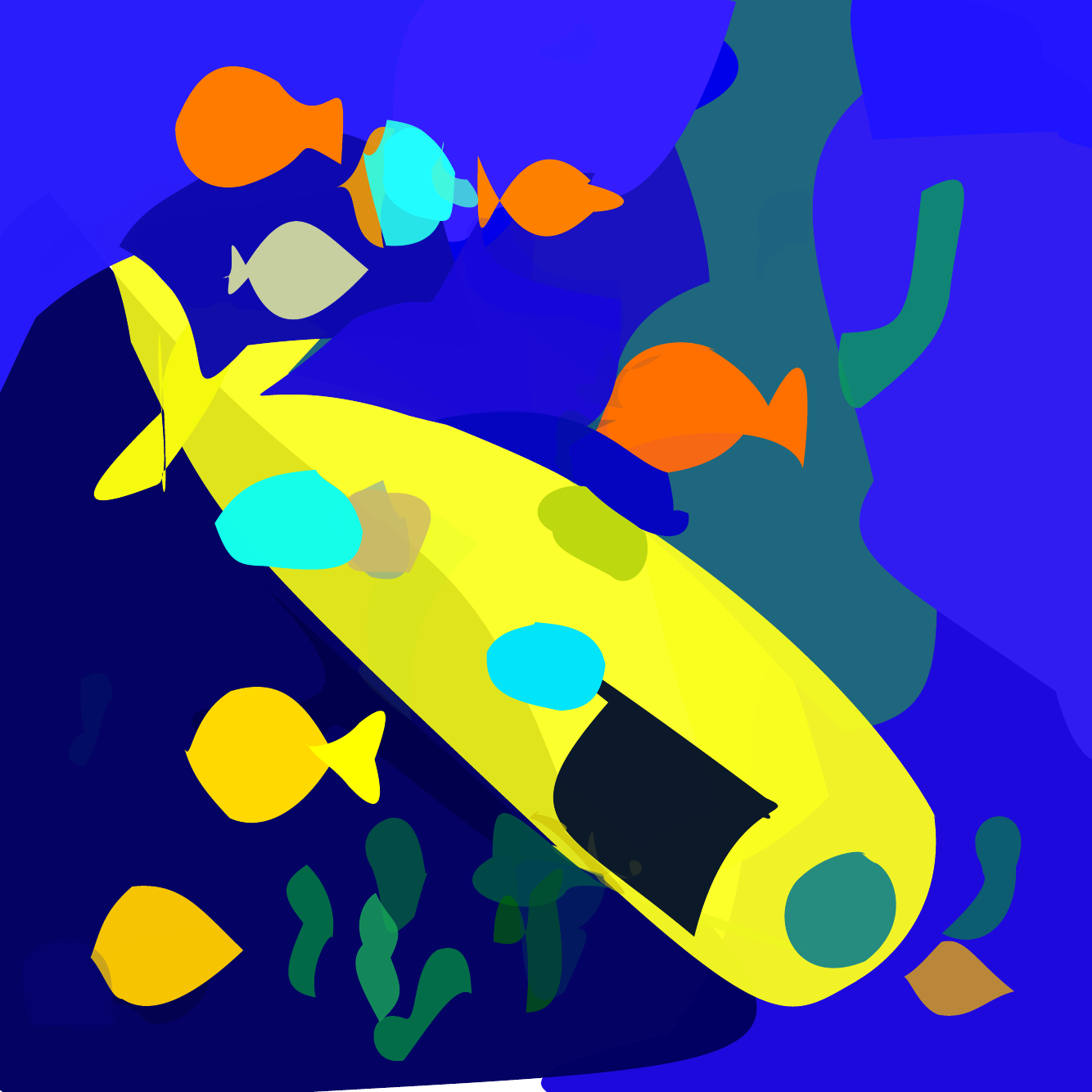}
    \tightcaption{\scriptsize{Underwater Submarine*}}
    \end{subfigure} &
    \begin{subfigure}[t]{\teaserwidthb}
    \includegraphics[width=\columnwidth]{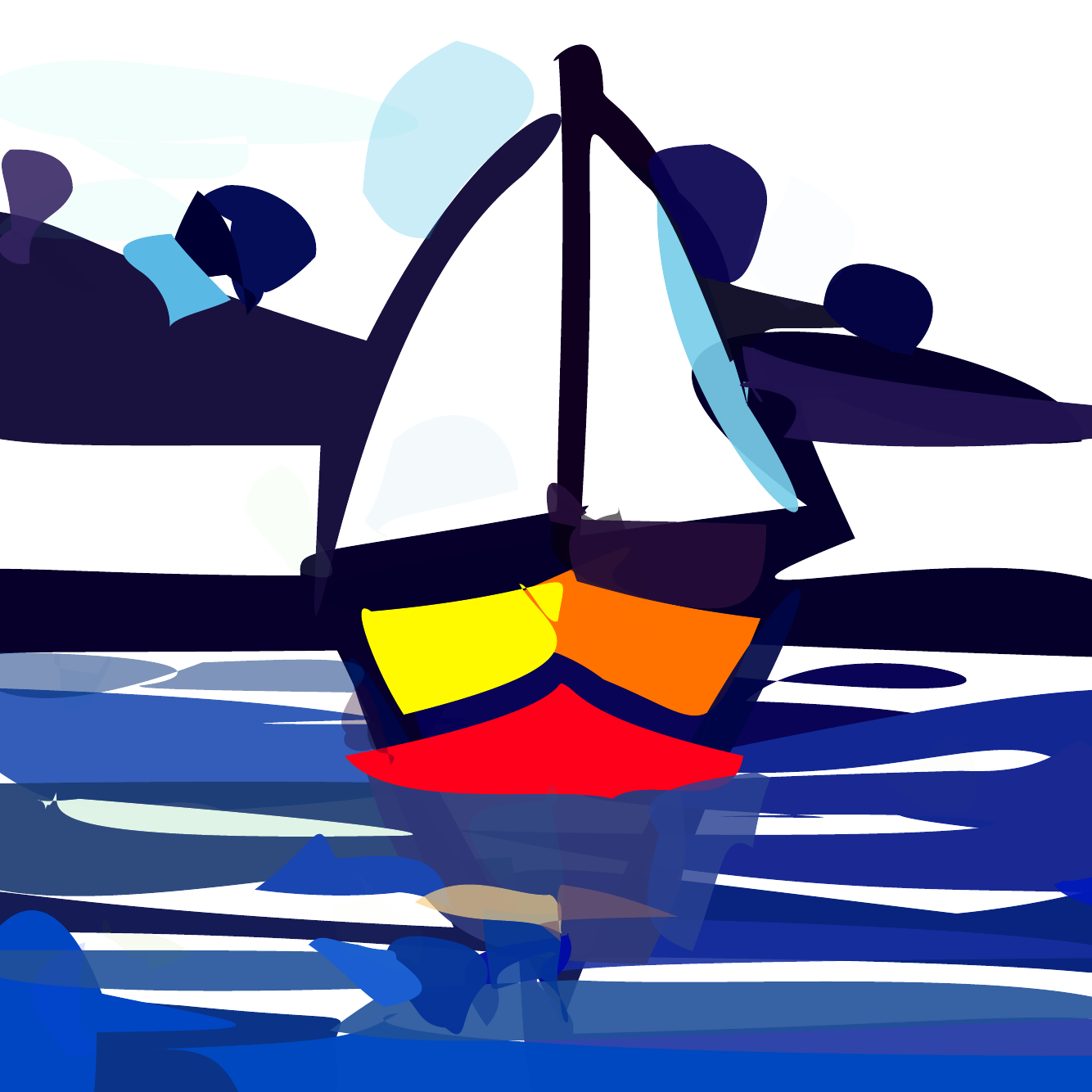}
    \tightcaption{\scriptsize{a boat*}}
    \end{subfigure} \\
    \begin{subfigure}[t]{\teaserwidthb}
    \includegraphics[width=\columnwidth]{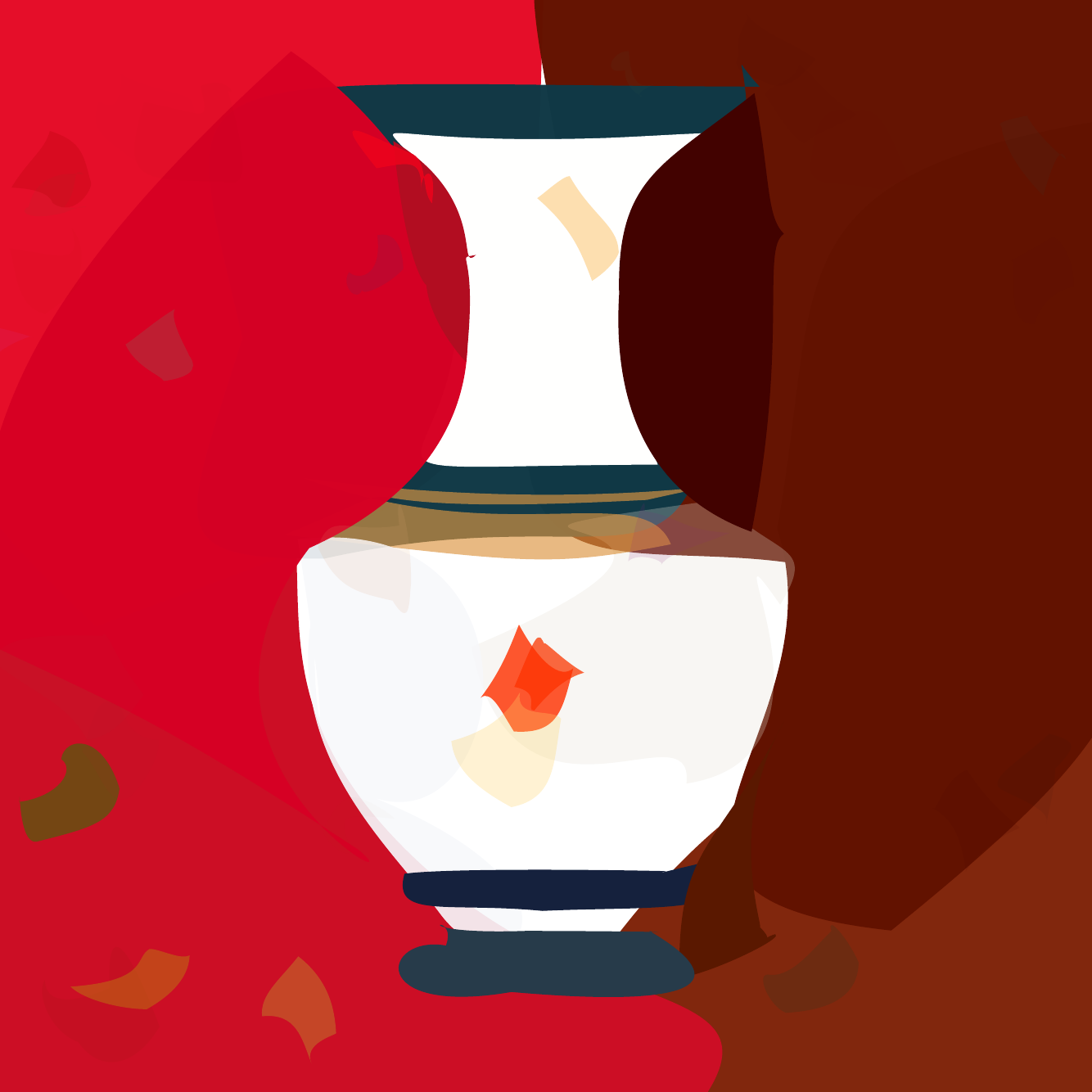}
    \tightcaption{\scriptsize{A photo of a Ming Dynasty vase on a leather topped table.*}}
    \end{subfigure} &
    \begin{subfigure}[t]{\teaserwidthb}
    \includegraphics[width=\columnwidth]{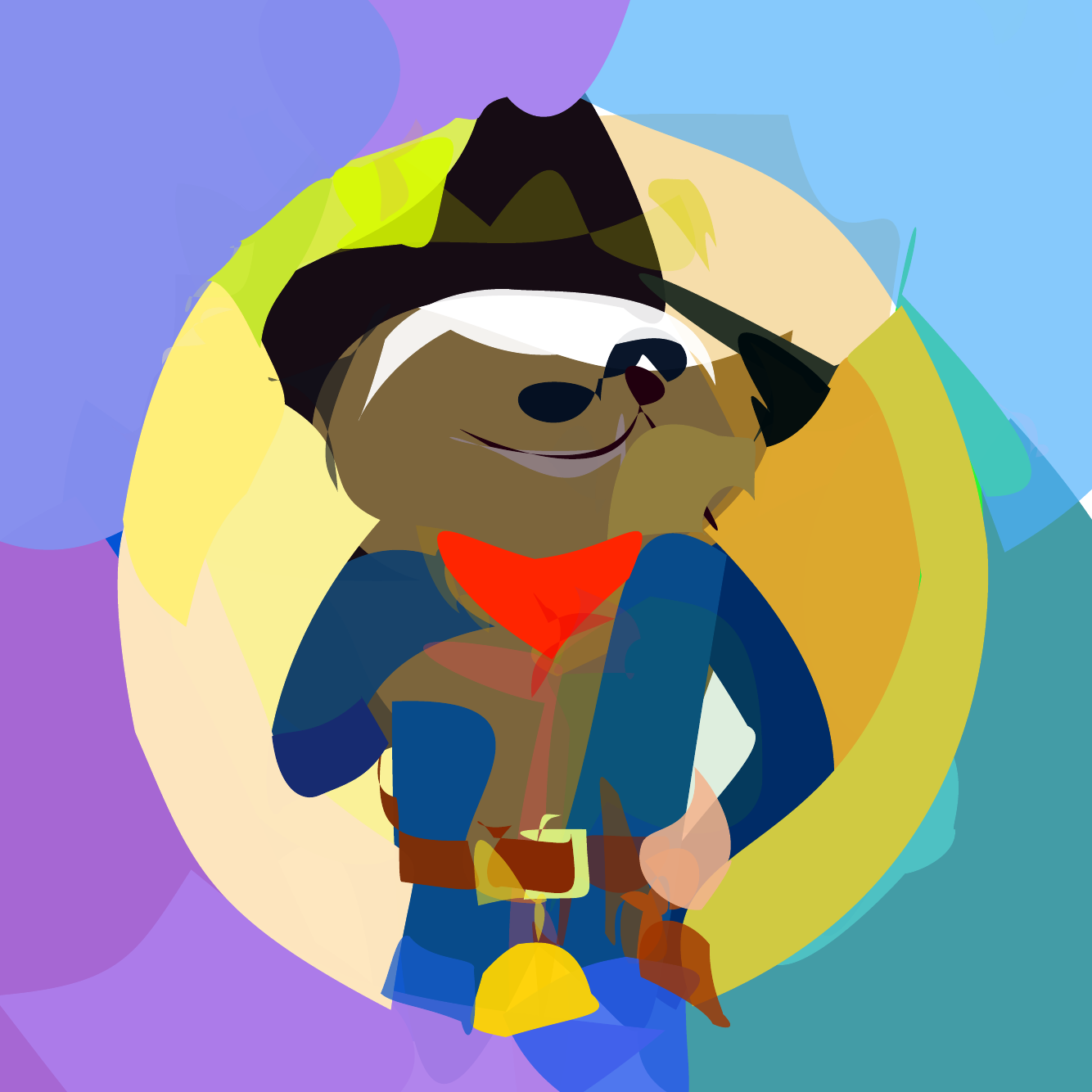}
    \tightcaption{\scriptsize{A smiling sloth wearing a leather jacket, a cowboy hat and a kilt.*}}
    \end{subfigure} &
    \begin{subfigure}[t]{\teaserwidthb}
    \includegraphics[width=\columnwidth]{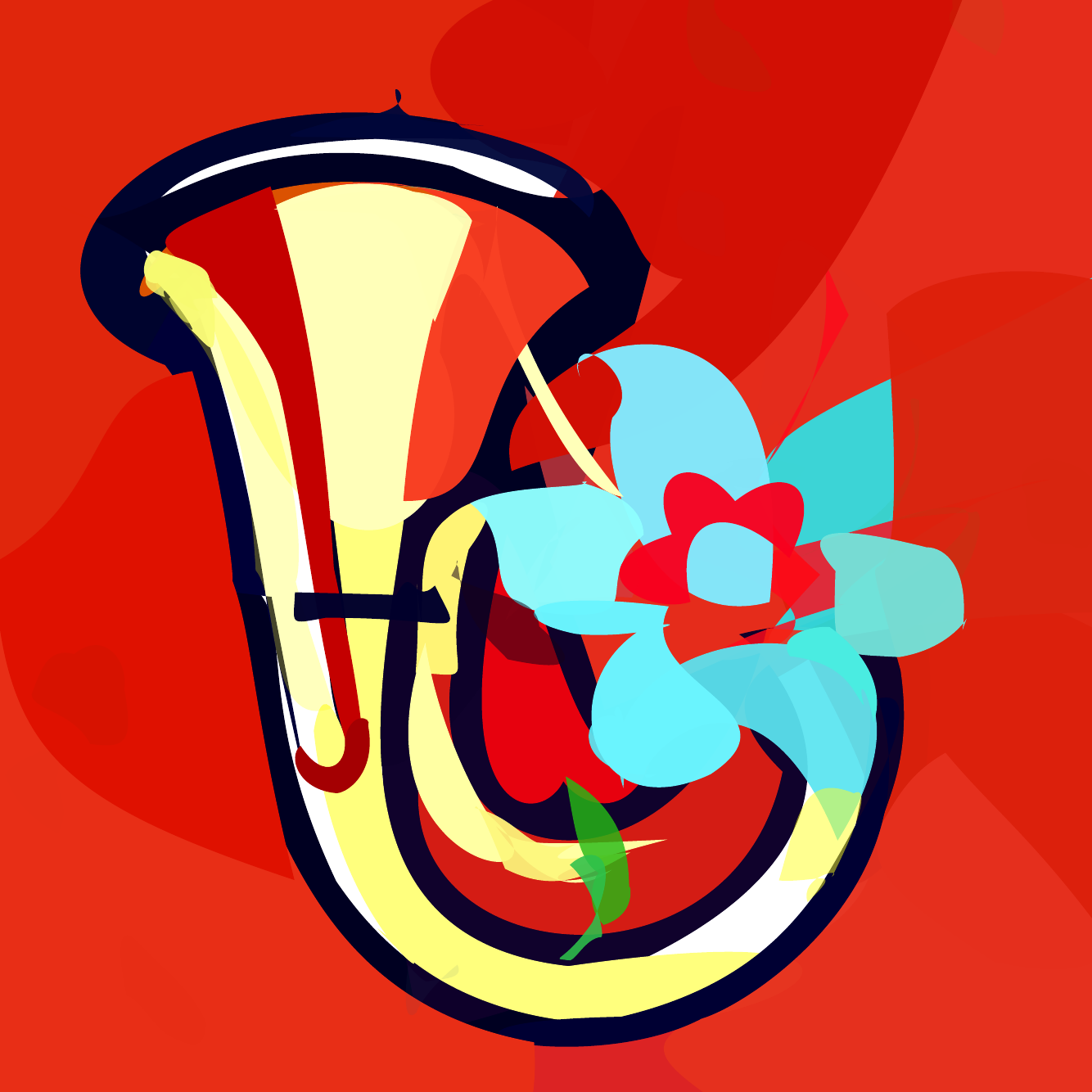}
    \tightcaption{\scriptsize{a tuba with red flowers protruding from its bell*}}
    \end{subfigure} &
    \begin{subfigure}[t]{\teaserwidthb}
    \includegraphics[width=\columnwidth]{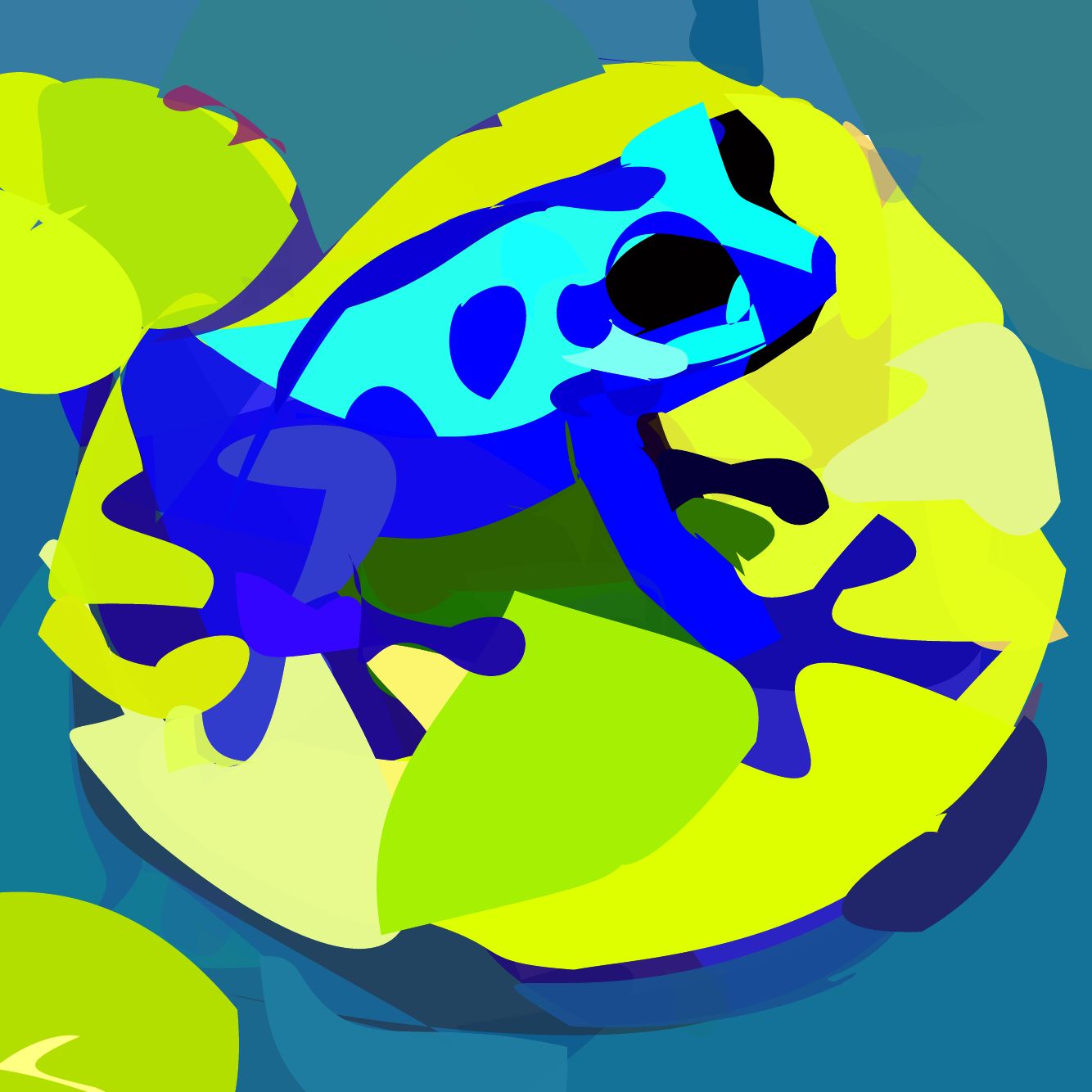}
    \tightcaption{\scriptsize{a blue poison dart frog sitting on a water lily*}}
    \end{subfigure} \\
    \begin{subfigure}[t]{\teaserwidthb}
    \includegraphics[width=\columnwidth]{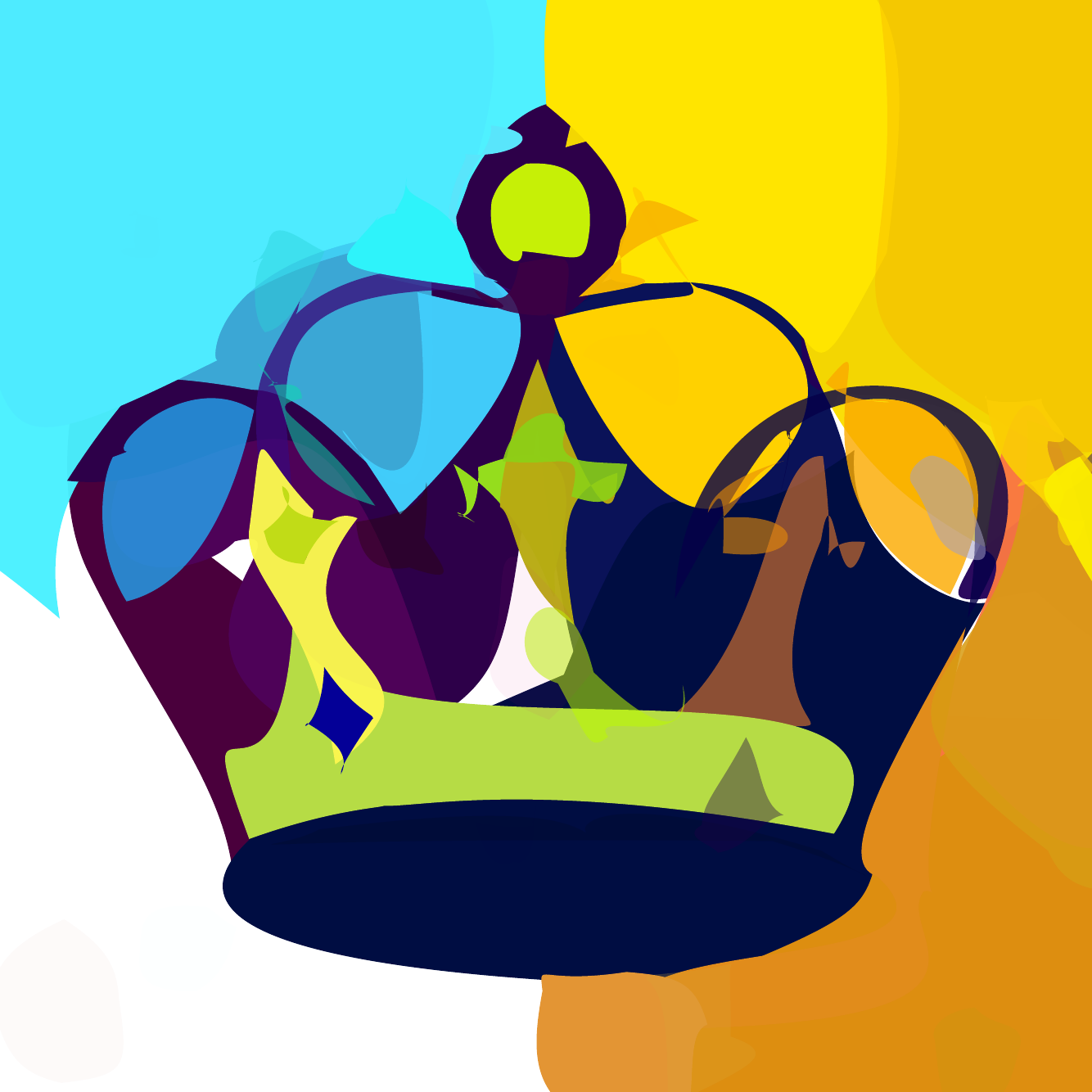}
    \tightcaption{\scriptsize{a crown*}}
    \end{subfigure} &
    \begin{subfigure}[t]{\teaserwidthb}
    \includegraphics[width=\columnwidth]{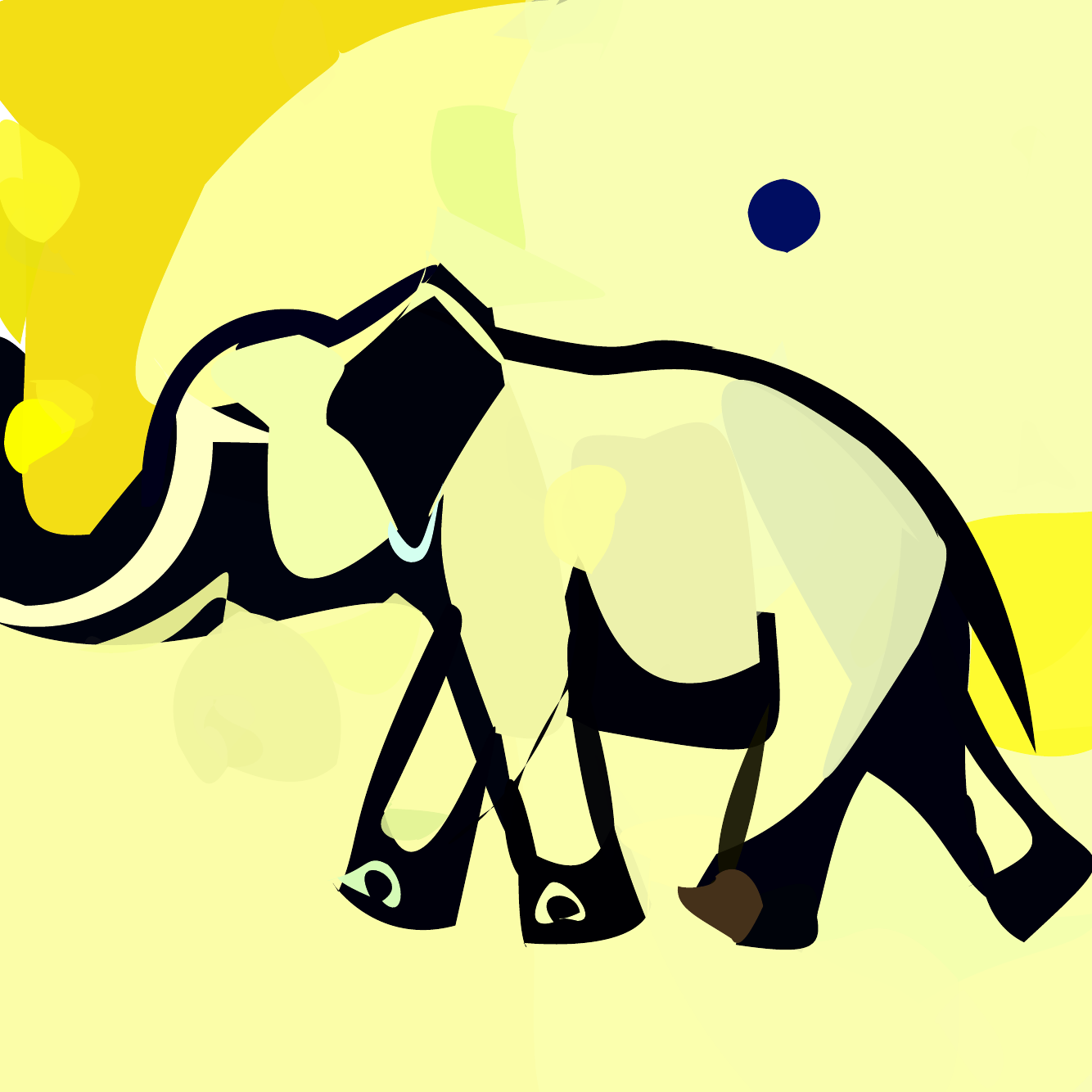}
    \tightcaption{\scriptsize{the silhouette of an elephant*}}
    \end{subfigure} &
    \begin{subfigure}[t]{\teaserwidthb}
    \includegraphics[width=\columnwidth]{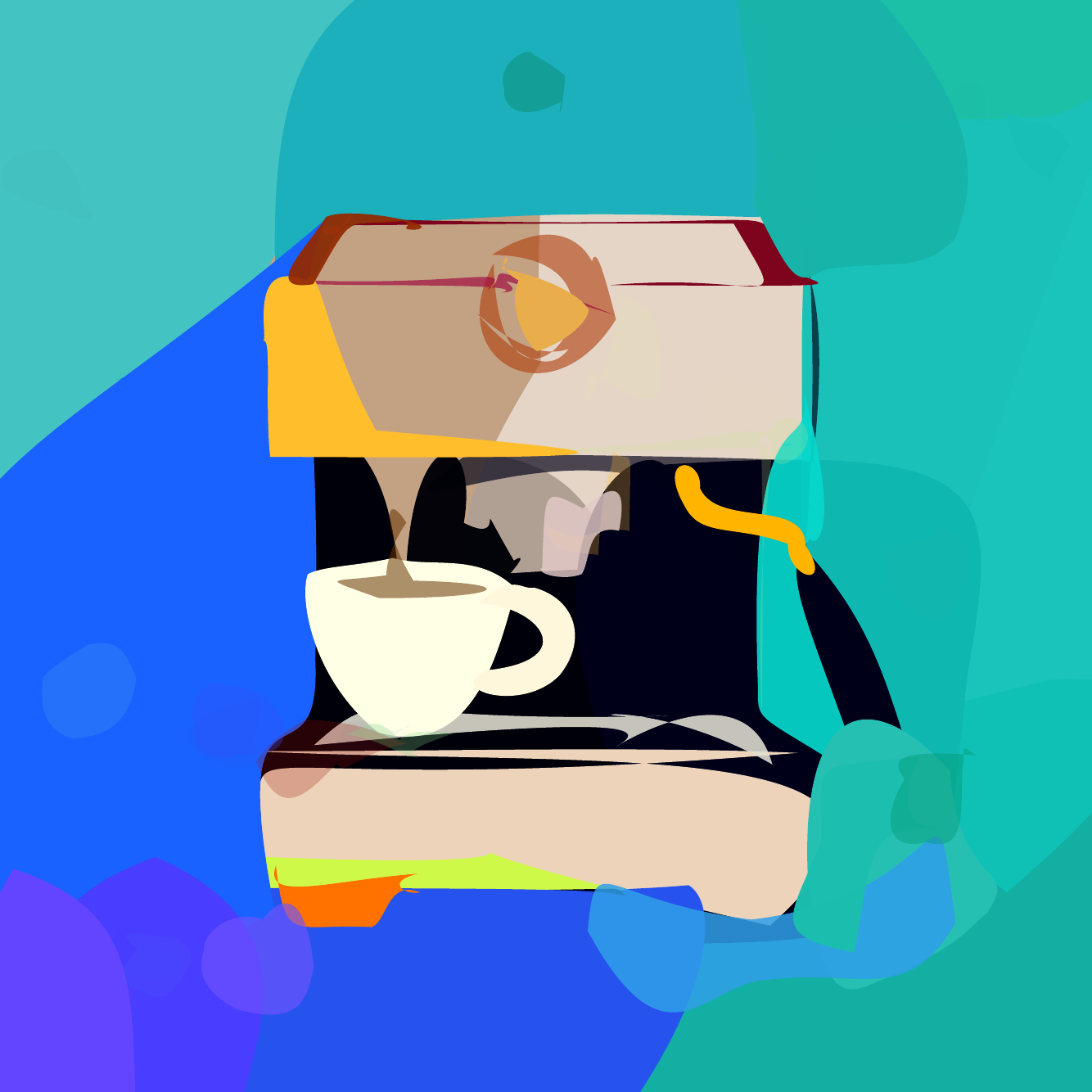}
    \tightcaption{\scriptsize{an espresso machine*}}
    \end{subfigure} &
    \begin{subfigure}[t]{\teaserwidthb}
    \includegraphics[width=\columnwidth]{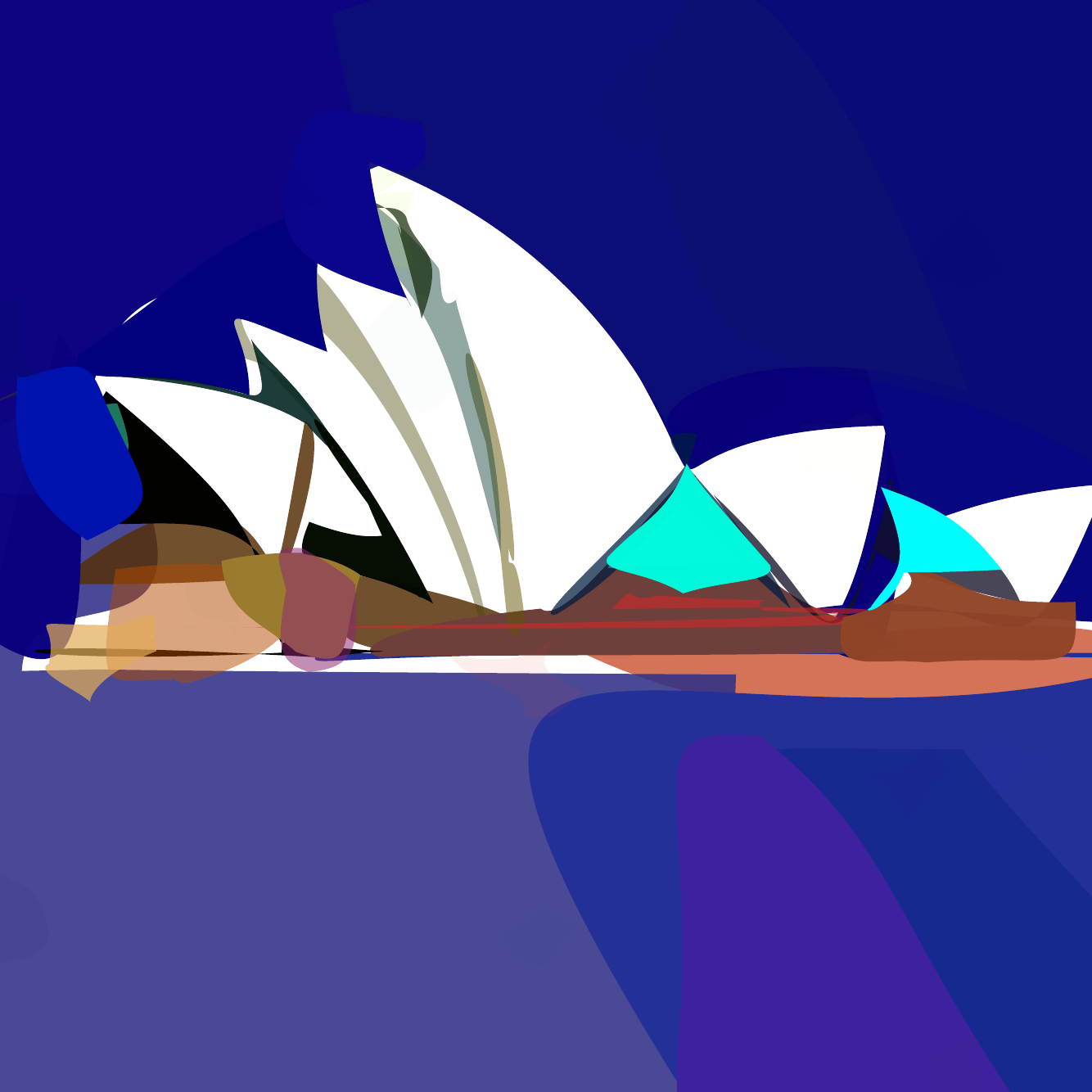}
    \tightcaption{\scriptsize{the Sydney Opera House*}}
    \end{subfigure}
\end{tabular}%
}
\resizebox{0.95\textwidth}{!}{%
\begin{tabular}{@{}c@{\quad}c@{\quad}c@{\quad}c@{\quad}c@{}}
    \begin{subfigure}[t]{\teaserwidthd}
    \includegraphics[width=\columnwidth]{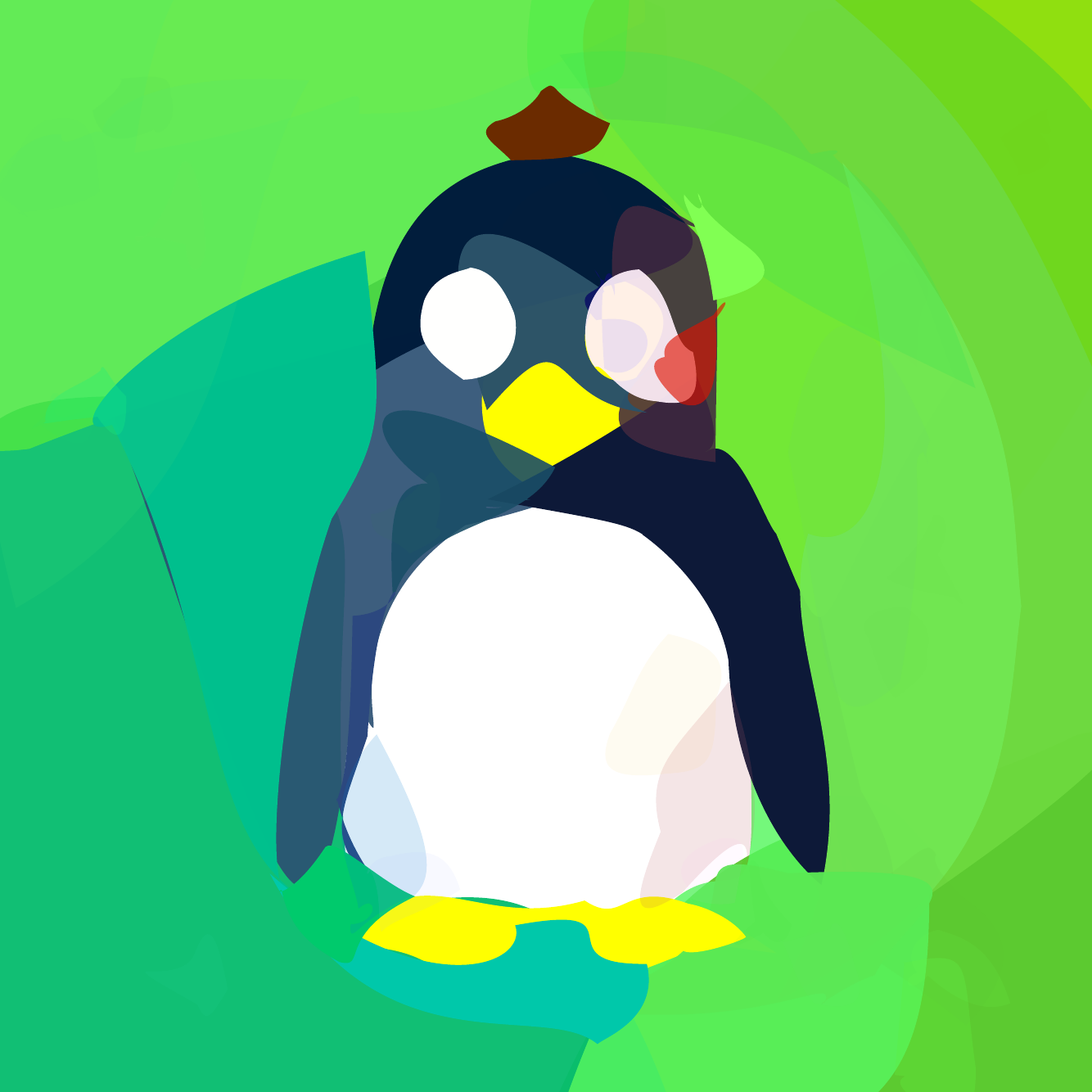}
    \tightcaption{\scriptsize{a baby penguin*}}
    \end{subfigure} &
    \begin{subfigure}[t]{\teaserwidthd}
    \includegraphics[width=\columnwidth]{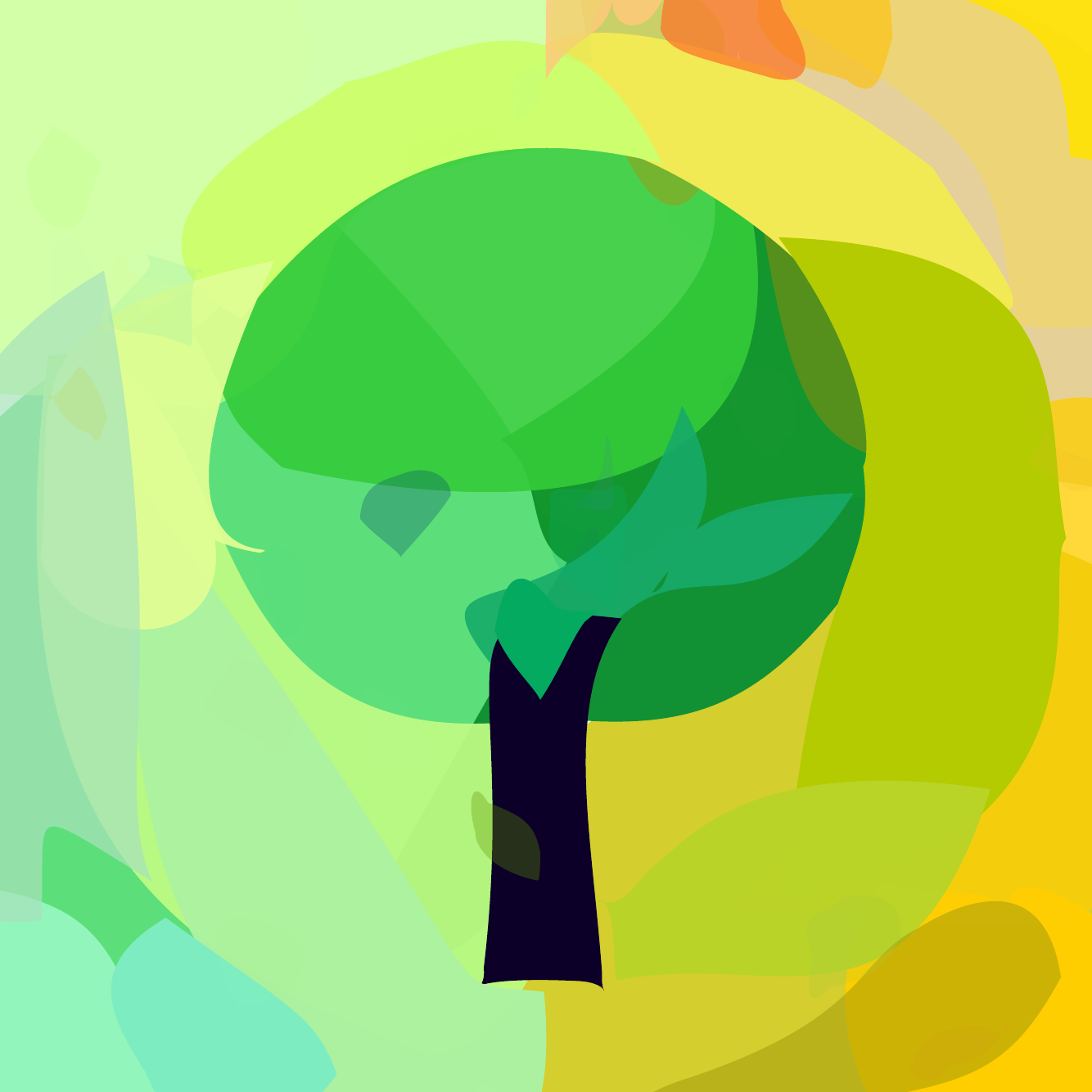}
    \tightcaption{\scriptsize{a tree*}}
    \end{subfigure} &
    \begin{subfigure}[t]{\teaserwidthd}
    \includegraphics[width=\columnwidth]{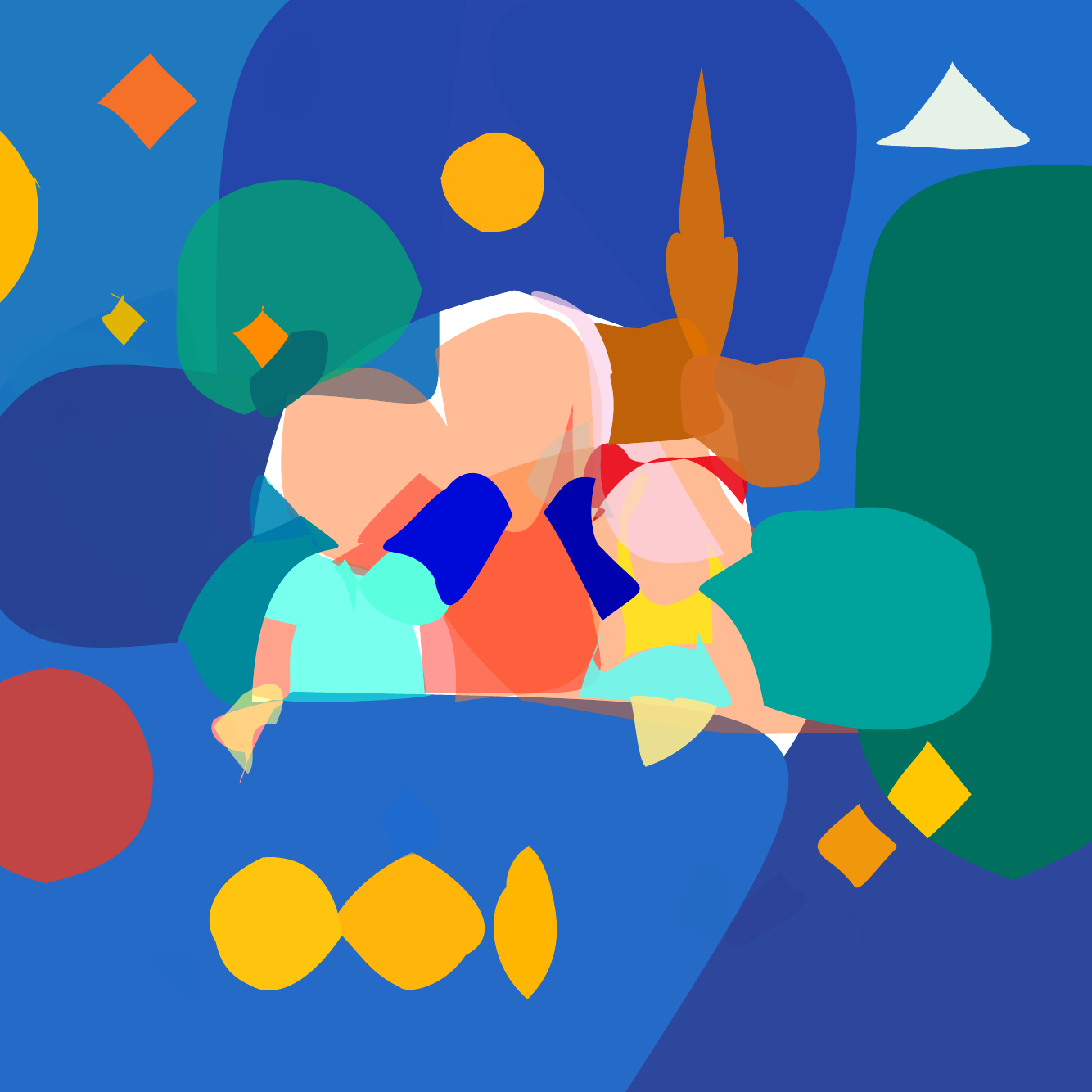}
    \tightcaption{\scriptsize{a family vacation to Walt Disney World*}}
    \end{subfigure} &
    \begin{subfigure}[t]{\teaserwidthd}
    \includegraphics[width=\columnwidth]{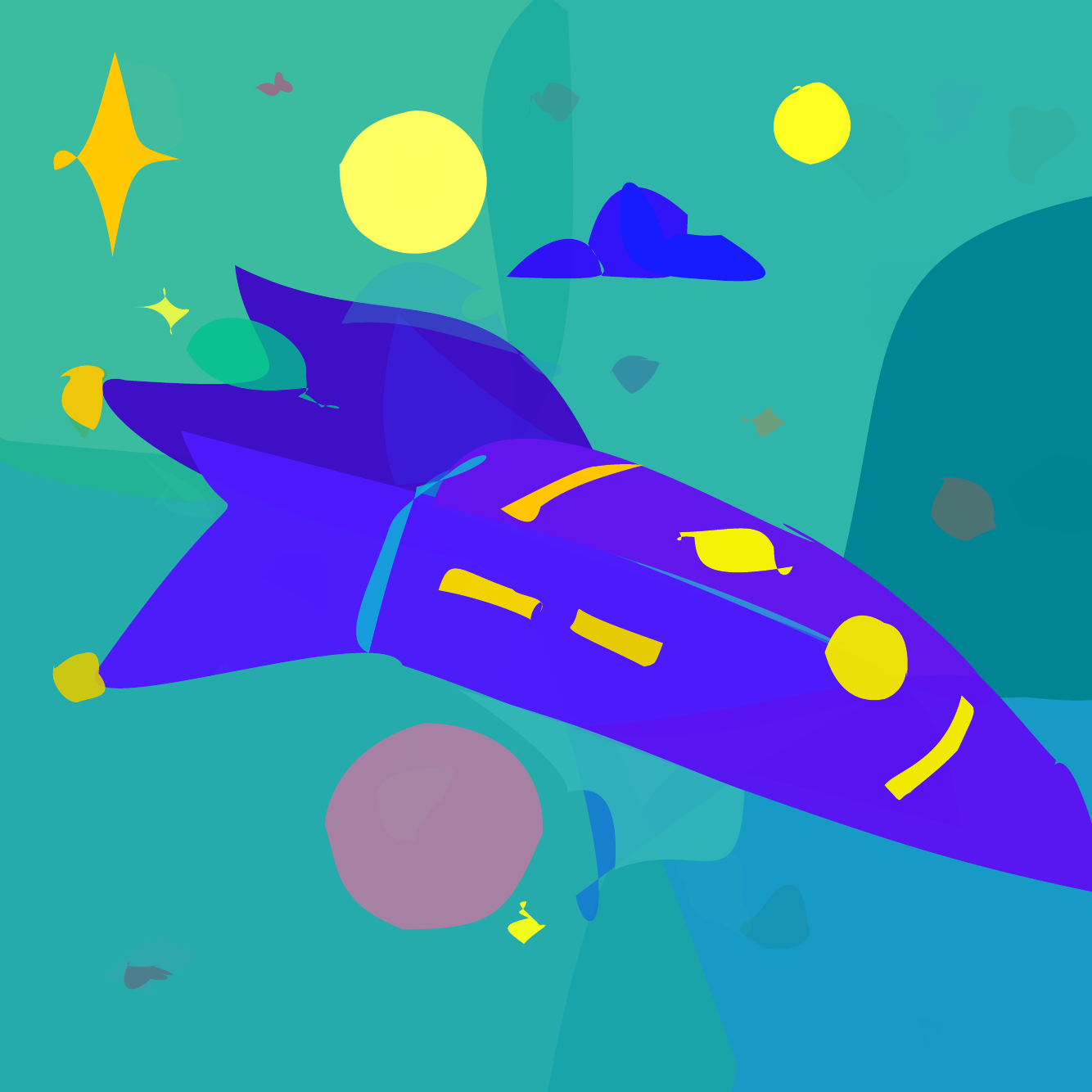}
    \tightcaption{\scriptsize{a spaceship flying in a starry night sky*}}
    \end{subfigure} &
    \begin{subfigure}[t]{\teaserwidthd}
    \includegraphics[width=\columnwidth]{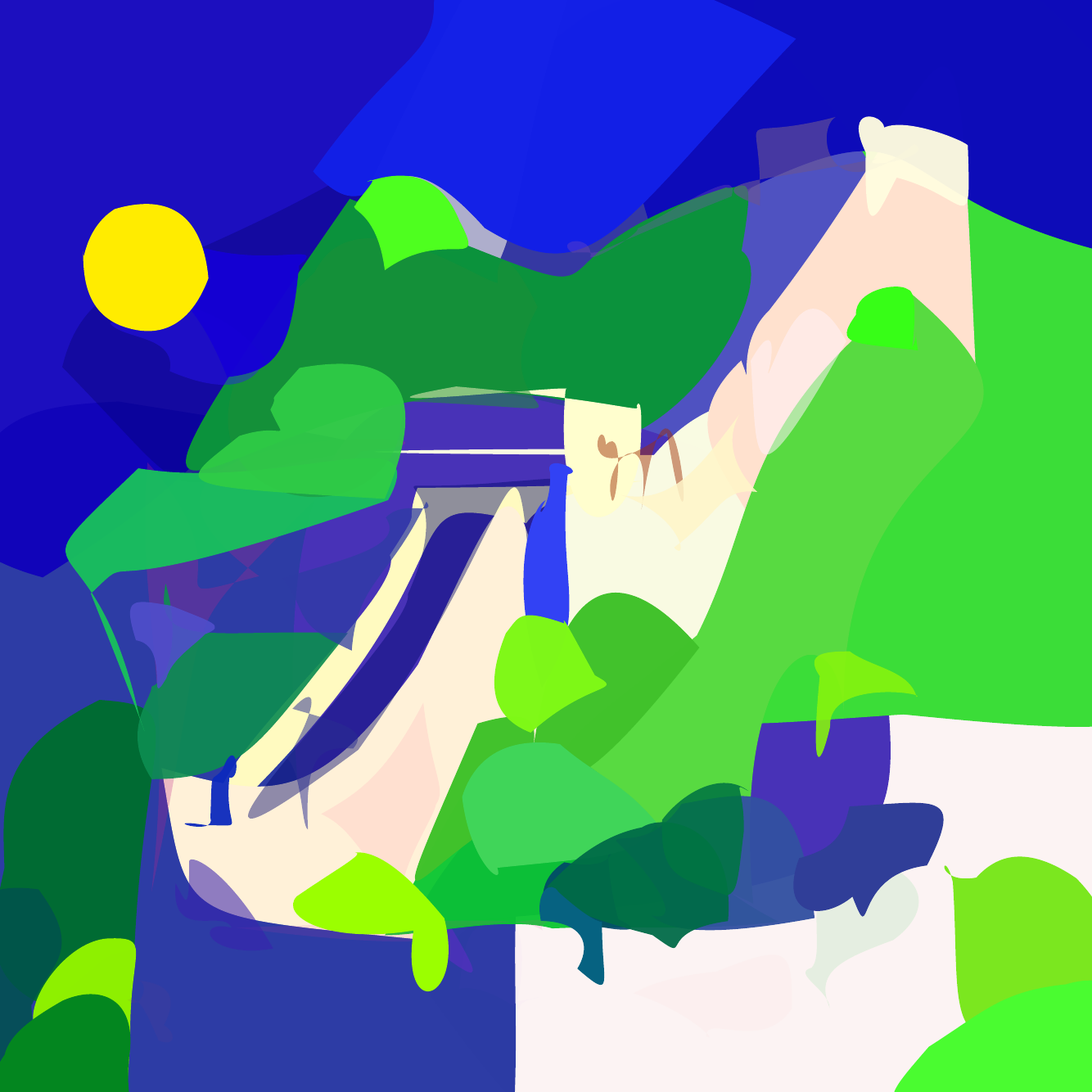}
    \tightcaption{\scriptsize{the Great Wall*}}
    \end{subfigure}
\end{tabular}%
}
\resizebox{0.95\textwidth}{!}{%
\begin{tabular}{@{}c@{\quad}c@{\quad}c@{\quad}c@{\quad}c@{\quad}c@{}}
    \begin{subfigure}[t]{\teaserwidthc}
    \includegraphics[width=\columnwidth]{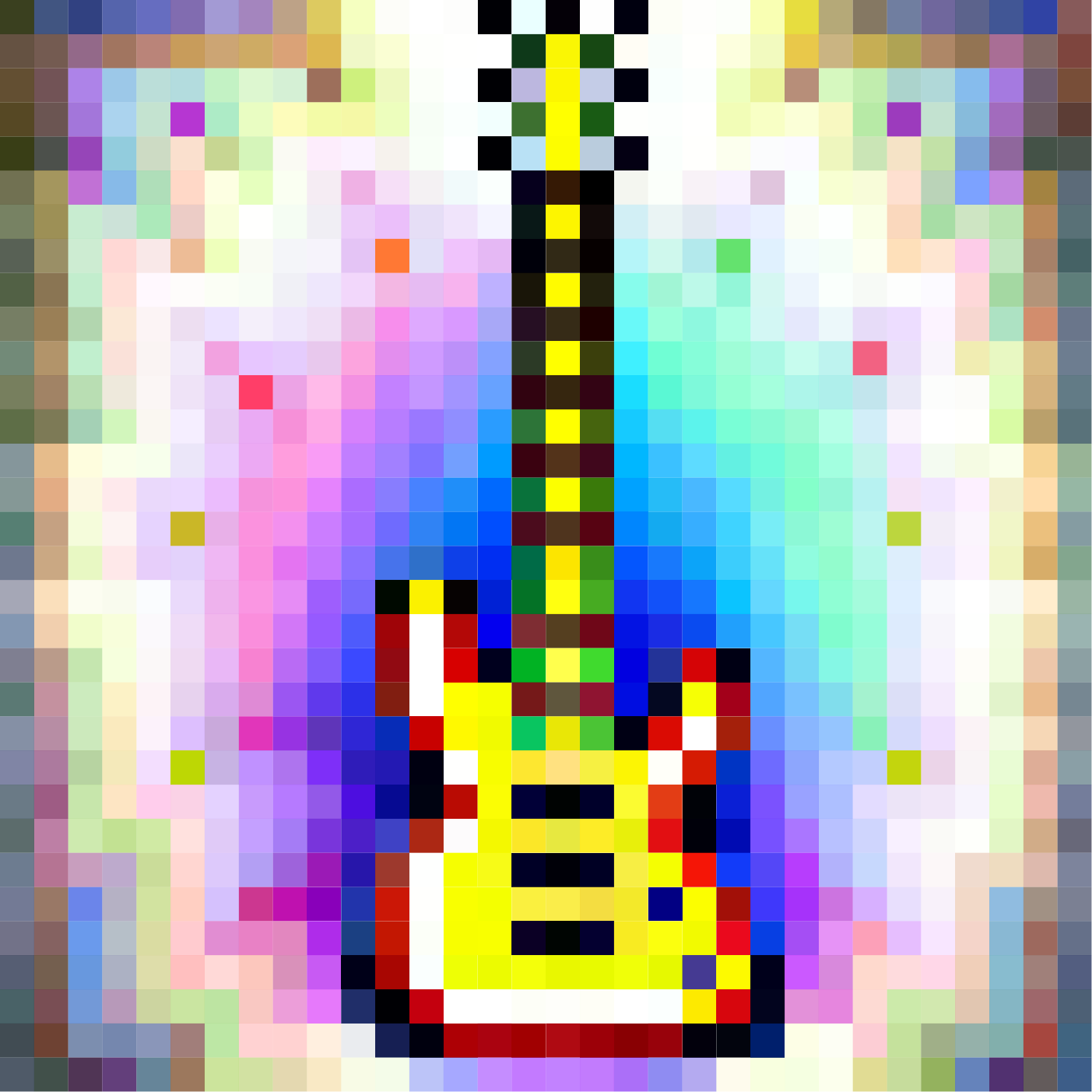}
    \tightcaption{\scriptsize{Electric guitar**}}
    \end{subfigure} &
    \begin{subfigure}[t]{\teaserwidthc}
    \includegraphics[width=\columnwidth]{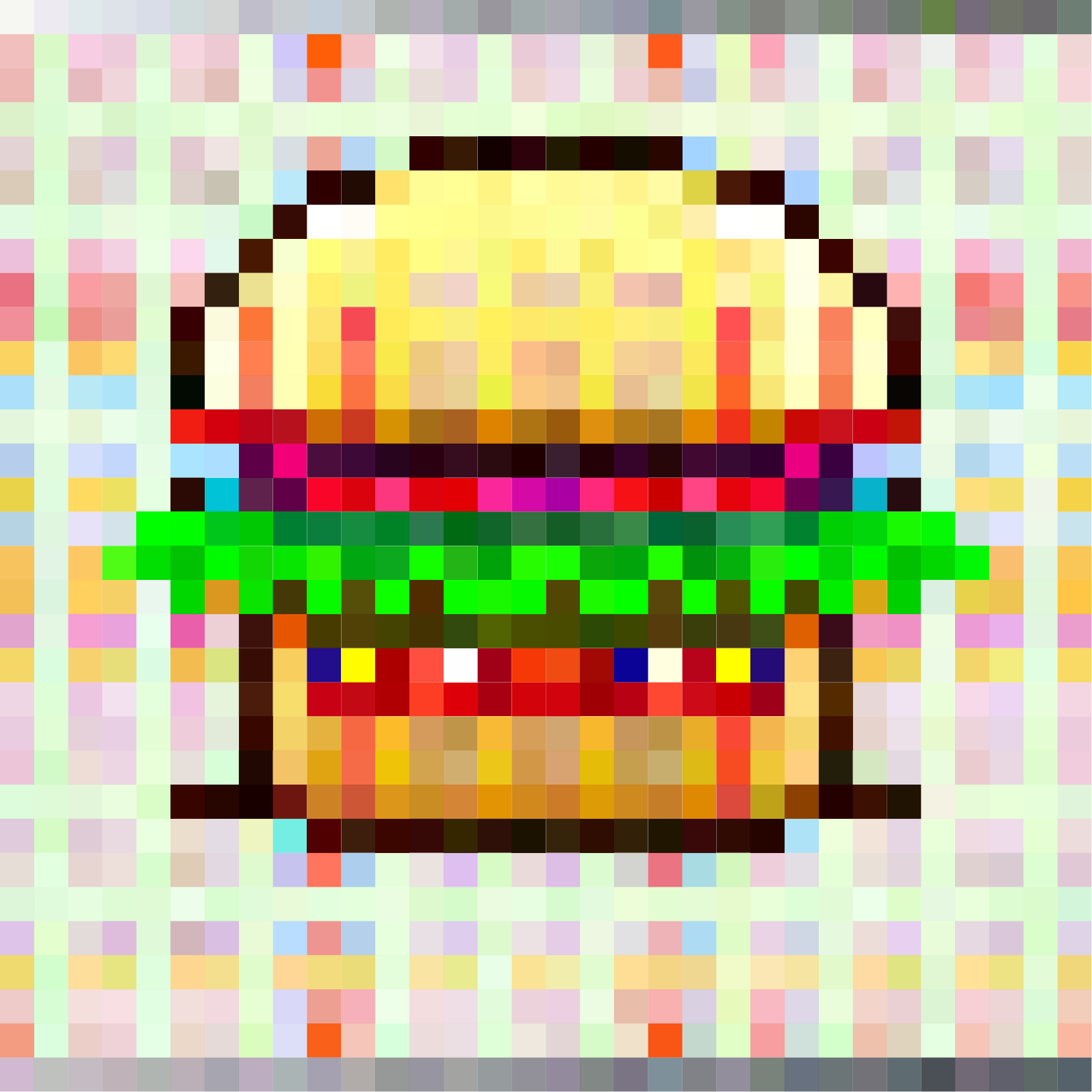}
    \tightcaption{\scriptsize{A delicious hamburger**}}
    \end{subfigure} &
    \begin{subfigure}[t]{\teaserwidthc}
    \includegraphics[width=\columnwidth]{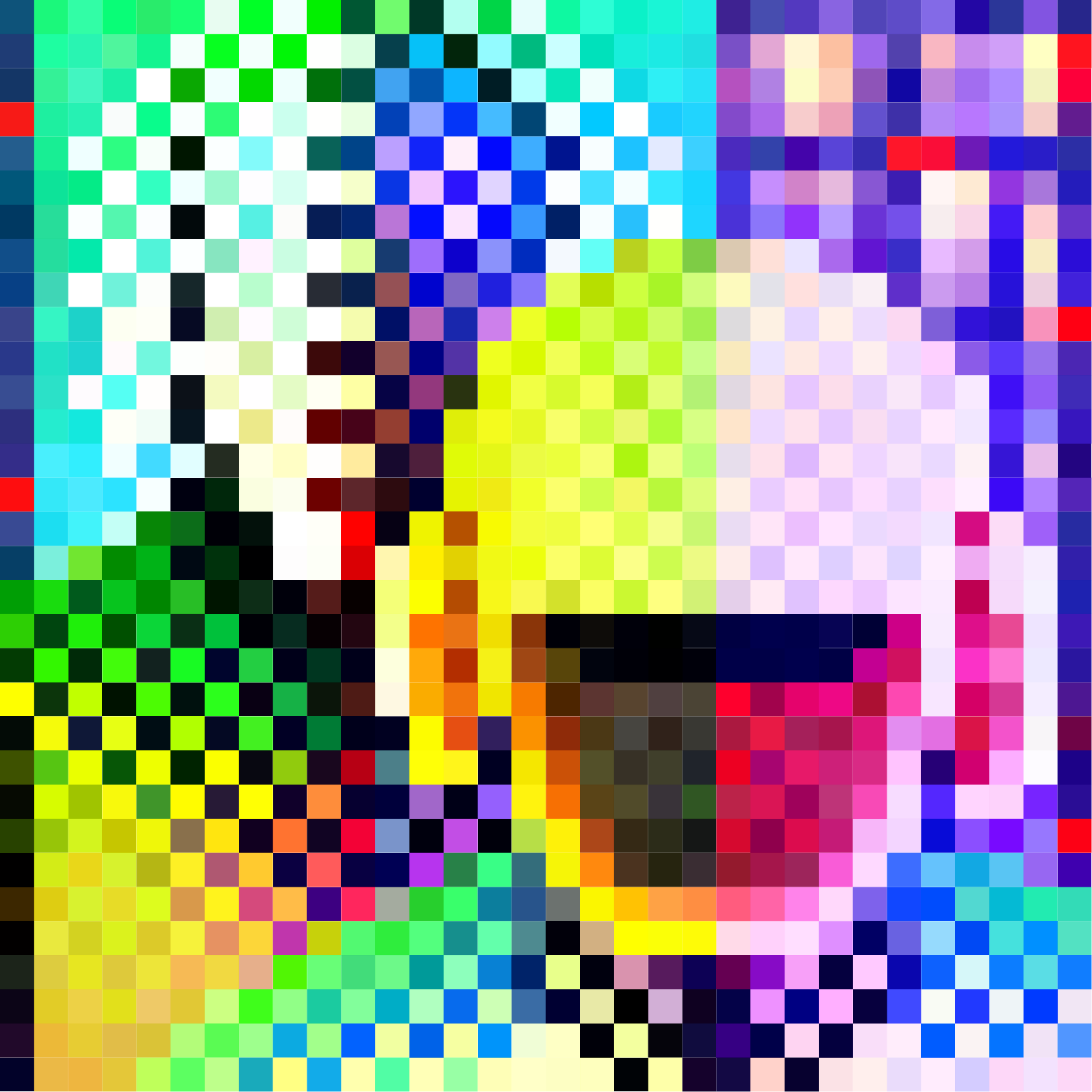}
    \tightcaption{\scriptsize{Daft Punk**}}
    \end{subfigure} &
    \begin{subfigure}[t]{\teaserwidthc}
    \includegraphics[width=\columnwidth]{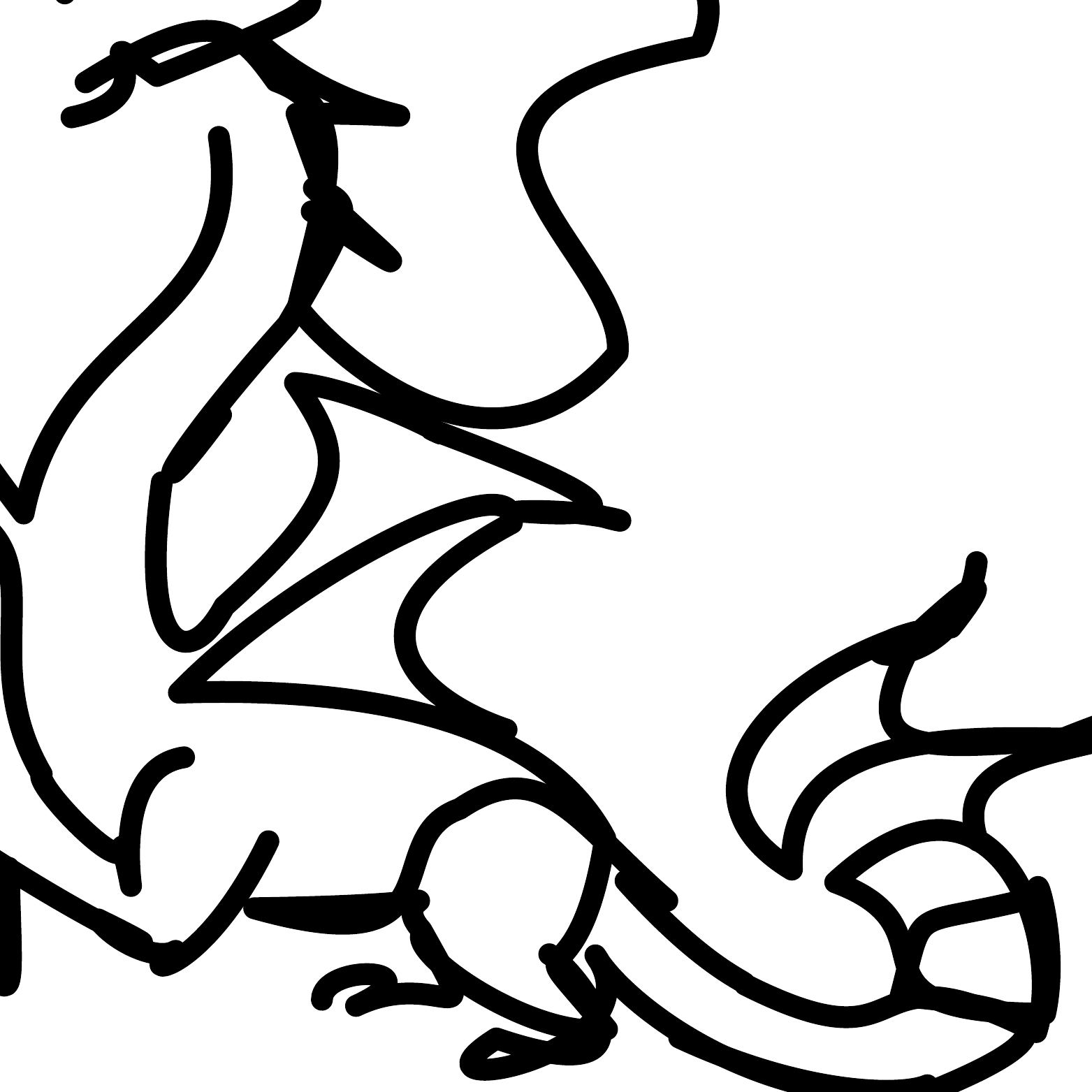}
    \tightcaption{\scriptsize{watercolor painting of a fire-breathing dragon$\dagger$}}
    \end{subfigure} &
    \begin{subfigure}[t]{\teaserwidthc}
    \includegraphics[width=\columnwidth]{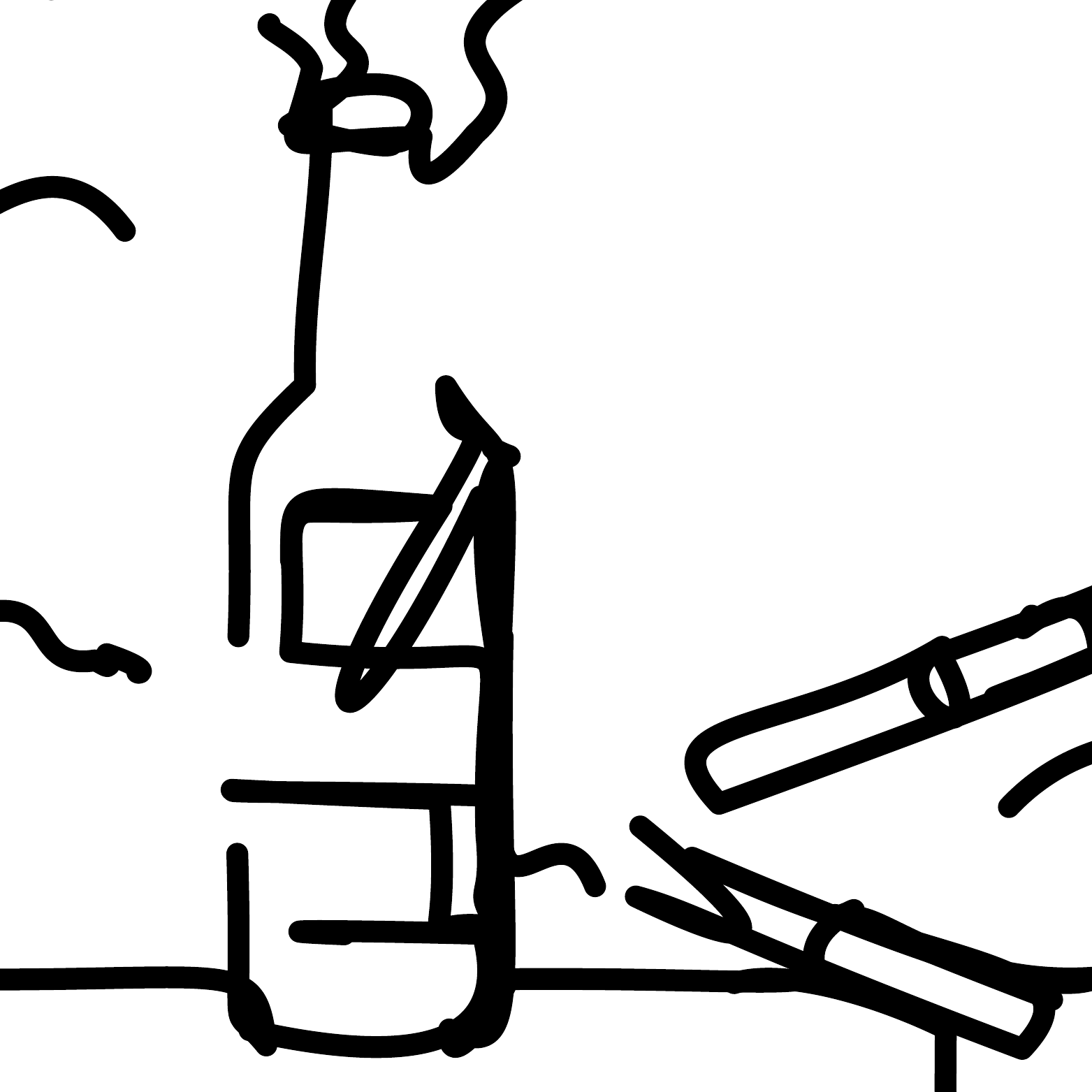}
    \tightcaption{\scriptsize{a bottle of beer next to an ashtray with a half-smoked cigarette$\dagger$}}
    \end{subfigure} &
    \begin{subfigure}[t]{\teaserwidthc}
    \includegraphics[width=\columnwidth]{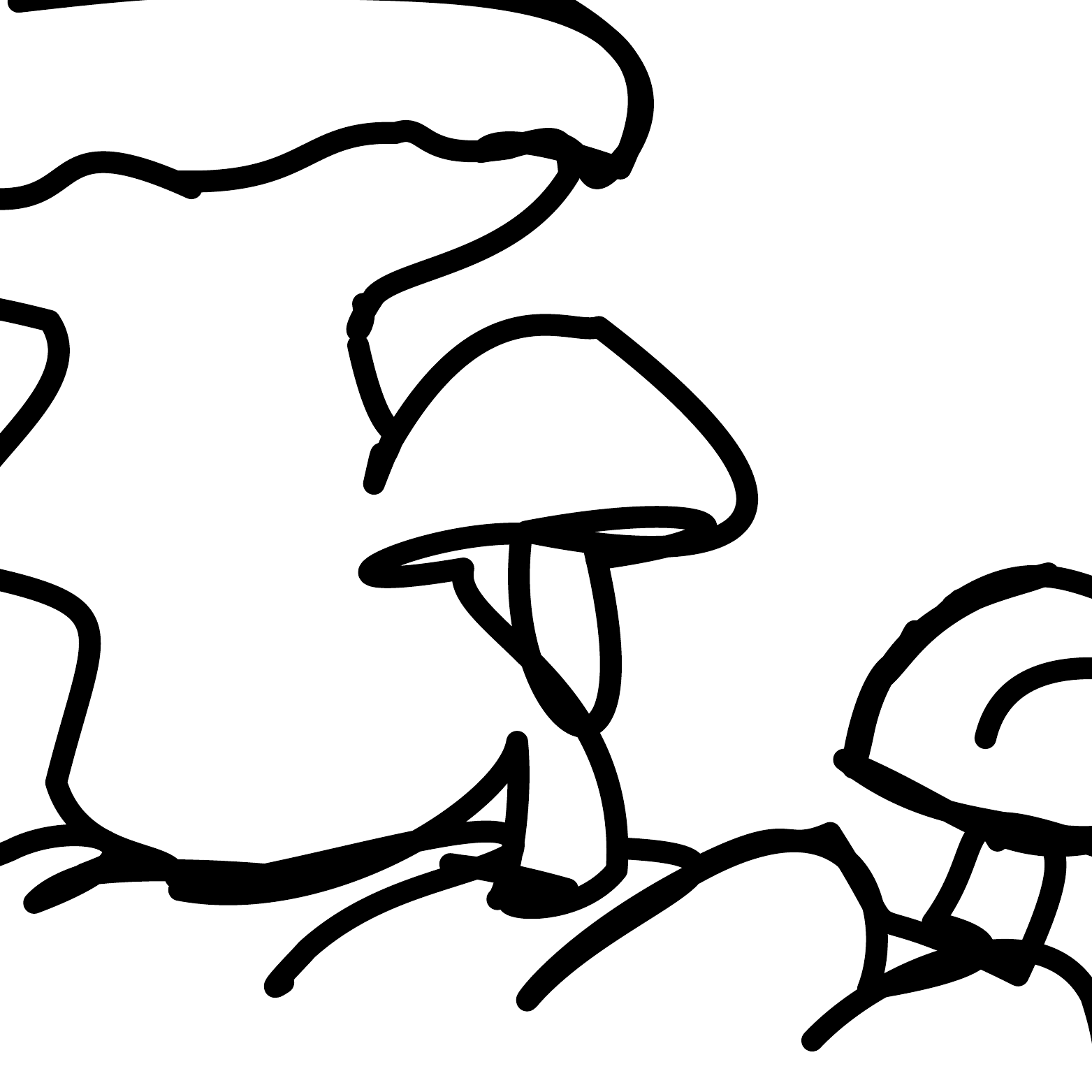}
    \tightcaption{\scriptsize{a brightly colored mushroom growing on a log$\dagger$}}
    \end{subfigure}
\end{tabular}%
}
\caption{
Given a caption, \ours{} generates abstract vector graphics in an SVG format. We use a pre-trained diffusion model trained only on rasterized images to guide a differentiable vector renderer. \ours{} supports diverse objects and styles. To select a style such as flat polygonal vector icons, abstract line drawings or pixel art, we constrain the vector representation to subset of possible primitive shapes and use different prompt modifiers to encourage an appropriate style:
* {\footnotesize ...minimal flat 2d vector icon. lineal color. on a white background. trending on artstation}, \,\,
** {\footnotesize ...pixel art. trending on artstation}, \,\, $\dagger${\footnotesize ...minimal 2d line drawing. trending on artstation.} Please see videos of the optimization process on our \href{https://ajayj.com/vectorfusion}{project webpage}. 
}
\label{fig:teaser}
\end{figure*}

Recently, large captioned datasets and breakthroughs in diffusion models have led to systems capable of generating diverse images from text including DALL-E 2~\cite{dalle2}, Imagen~\cite{imagen} and Latent Diffusion~\cite{rombach2021highresolution}. However, the vast majority of images available in web-scale datasets are rasterized, expressed at a finite resolution with no decomposition into primitive parts nor layers. For this reason, existing diffusion models can only generate raster images. 
In theory, diffusion models could be trained to directly model SVGs, but would need specialized architectures for variable-length hierarchical sequences, and significant data collection work.

How can we use diffusion models pretrained on pixels to generate high-quality vector graphics? In this work, we provide a method for generating high quality abstract vector graphics from text captions, shown in Fig.~\ref{fig:sd_comparison}.

We start by evaluating a two phase text-to-image and image-to-vector baseline: generating a raster image with a pretrained diffusion model, then vectorizing it. Traditionally, designers manually convert simple rasterized images into a vector format by tracing shapes. Some ML-based tools~\cite{xu2022live} can automatically approximate a raster image with an SVG. Unfortunately, we find that text-to-image diffusion models frequently produce complex images that are hard to represent with simple vectors, or are incoherent with the caption (Fig~\ref{fig:sd_comparison}, Stable Diffusion + LIVE). Even with a good pixel sample, automated conversion loses details.

To improve quality of the SVG and coherence with the caption, we incorporate the pretrained text-to-image diffusion model in an optimization loop. Our approach, \ours{}, combines a differentiable vector graphics renderer~\cite{Li:2020:DVG} and a recently proposed \textit{score distillation sampling} (SDS) loss~\cite{poole2022dreamfusion} to iteratively refine shape parameters. Intuitively, score distillation converts diffusion sampling into an optimization problem that allows the image to be represented by an arbitrary differentiable function. In our case, the differentiable function is the forward rasterization process, and the diffusion model provides a signal for improving the raster. To adapt SDS to text-to-SVG synthesis, we make the following contributions:
\begin{itemize}[noitemsep,topsep=0pt]
    \item We extend score distillation sampling to open source latent space diffusion models like Stable Diffusion,
    \item improve efficiency and quality by initializing near a raster image sample,
    \item propose SVG-specific regularization including path reinitialization,
    \item and evaluate different sets of shape primitives and their impact on style.
\end{itemize}
In experiments, \ours{} generates iconography, pixel art and line drawings from diverse captions. \ours{} also achieves greater quality than CLIP-based approaches that transfer a discriminative vision-language representation.

\section{Related Work}
\label{sec:relatedwork}

A few works have used pretrained vision-language models to guide vector graphic generation. VectorAscent~\cite{jain21vector} and CLIPDraw~\cite{frans21clipdraw} optimize CLIP's image-text similarity metric~\cite{clip} to generate vector graphics from text prompts, with a procedure similar to DeepDream~\cite{mordvintsev2018differentiable} and CLIP feature visualization~\cite{goh2021multimodal}. StyleCLIPDraw~\cite{styleclipdraw} extends CLIPDraw to condition on images with an auxiliary style loss with a pretrained VGG16~\cite{https://doi.org/10.48550/arxiv.1409.1556} model. Arnheim \cite{fernando2021genart} parameterizes SVG paths with a neural network, and CLIP-CLOP \cite{mirowski2022clip} uses an evolutionary approach to create image collages. Though we also use pretrained vision-language models, we use a generative model, Stable Diffusion~\cite{rombach2021highresolution} rather than a discriminative model. 

Recent work has shown the success of text-to-image generation. DALL-E 2~\cite{dalle2} learns an image diffusion model conditioned on CLIP's text embeddings. Our work uses Stable Diffusion~\cite{rombach2021highresolution} (SD), a text-to-image latent diffusion model. While these models produce high-fidelity images, they cannot be directly transformed into vector graphics. 

A number of works generate vector graphics from input images. We extend the work of Layer-wise Image Vectorization (LIVE)~\cite{xu2022live}, which iteratively optimizes closed Bézier paths with a differentiable rasterizer, DiffVG~\cite{Li:2020:DVG}. 

We also take inspiration from inverse graphics with diffusion models. Diffusion models have been used in zero-shot for image-to-image tasks like inpainting~\cite{repaint}. DDPM-PnP~\cite{ddpmpnp} uses diffusion models as priors for conditional image generation, segmentation, and more. DreamFusion~\cite{poole2022dreamfusion} uses 2D diffusion as an image prior for text-to-3D synthesis with a more efficient and high-fidelity loss than DDPM-PnP, discussed in Section~\ref{sec:sds}. Following~\cite{poole2022dreamfusion}, we use diffusion models as transferable priors for vector graphics. Concurrent work~\cite{metzer2022latentnerf} also adapts the SDS loss for latent-space diffusion models.

\section{Background}
\label{sec:background}

\subsection{Vector representation and rendering pipeline}

Vector graphics are composed of primitives. For our work, we use paths of segments delineated by control points. We configure the control point positions, shape fill color, stroke width and stroke color. Most of our experiments use closed B\'ezier curves. Different artistic styles are accomplished with other primitives, such as square shapes for pixel-art synthesis and unclosed B\'ezier curves for line art.

To render to pixel-based formats, we rasterize the primitives. While many primitives would be needed to express a realistic photogaph, even a small number can be combined into recognizable, visually pleasing objects. We use DiffVG~\cite{Li:2020:DVG}, a differentiable rasterizer that can compute the gradient of the rendered image with respect to the parameters of the SVG paths. Many works, such as LIVE~\cite{xu2022live}, use DiffVG to vectorize images, though such transformations are lossy.

\subsection{Diffusion models}

Diffusion models are a flexible class of likelihood-based generative models that learn a distribution by denoising. A diffusion model generates data by learning to gradually map samples from a known prior like a Gaussian toward the data distribution. During training, a diffusion model optimizes a variational bound on the likelihood of real data samples~\cite{pmlr-v37-sohl-dickstein15}, similar to a variational autoencoder~\cite{Kingma2014AutoEncodingVB}. This bound reduces to a weighted mixture of denoising objectives~\cite{ddpm}:
\begin{equation}
    \mathcal{L}_\text{DDPM}(\phi, \x) = \mathbb{E}_{t, \epsilon} \left[ w(t) \| \epsilon_\phi(\alpha_t \x + \sigma_t \epsilon) - \epsilon \right \|^2_2]
\end{equation}
where $\x$ is a real data sample and $t\in \{1, 2, \ldots T\}$ is a uniformly sampled timestep scalar that indexes noise schedules $\alpha_t, \sigma_t$~\cite{kingma2021on}. $\epsilon$ is noise of the same dimension as the image sampled from the known Gaussian prior. Noise is added by interpolation to preserve variance. $\epsilon_\phi$ is a learned denoising autoencoder that predicts the noise content of its input. For images, $\epsilon_\phi$ is commonly a U-Net~\cite{unet, ddpm}, and the weighting function $w(t)=1$~\cite{ddpm}. Denoising diffusion models can be trained to predict any linear combination of $\x$ and $\epsilon$, such as the clean, denoised image $\x$, though an $\epsilon$ parameterization is simple and stable.

At test time, a sampler starts with a draw from the prior $\x_T \sim \mathcal{N}(0, 1)$, then iteratively applies the denoiser to update the sample while decaying the noise level $t$ to 0. For example, DDIM~\cite{ddim} samples with the update:
\begin{align}
    \hat{\x} &= (\x_t - \sigma_t \epsilon_\phi(\x_t)) / \alpha_t,~~&\text{Predict clean image} \nonumber \\
    \x_{t-1} &= \alpha_{t-1} \hat{\x} + \sigma_{t-1} \epsilon_\phi(\x_t)~~&\text{Add back noise} \label{eq:ddim}
\end{align}
For text-to-image generation, the U-Net is conditioned on the caption $y$, $\epsilon_\phi(\x, y)$, usually via cross-attention layers and pooling of the features of a language model~\cite{Nichol2022GLIDETP}. However, conditional diffusion models can produce results incoherent with the caption since datasets are weakly labeled and likelihood-based models try to explain all possible images. To increase the usage of a label or caption, classifier-free guidance~\cite{classifierfree} superconditions the model by scaling up conditional model outputs and guiding away from a generic unconditional prior that drops $y$:
\begin{equation}
    \hat{\epsilon}_\phi(\x, y) = (1 + \omega) * \epsilon_\phi(\x, y) - \omega * \epsilon_\phi(\x)
\end{equation}
CFG significantly improves coherence with a caption at the cost of an additional unconditional forward pass per step.

\begin{figure*}[t]
\begin{minipage}{.7\linewidth}
  \centering
    \includegraphics[width=\linewidth,trim={6cm 0 0 0},clip]{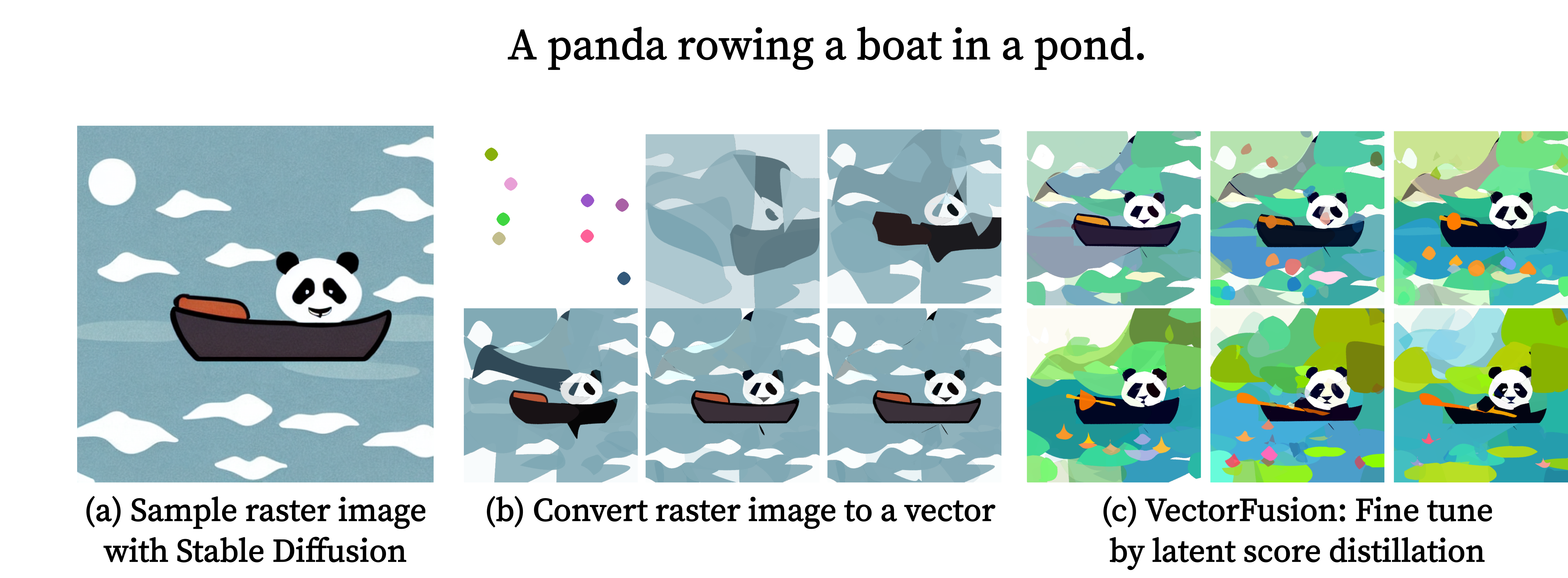}
    \captionof{figure}{
\ours{} generates SVGs in three stages. \textbf{(a)} First, we sample a rasterized image from a text-to-image diffusion model like Stable Diffusion with prompt engineering for iconographic aesthetics. \textbf{(b)} Since this image is finite resolution, we approximate it by optimizing randomly initialized vector paths with an L2 loss. The loss is backpropagated through DiffVG, a differentiable vector graphics renderer, to tune path coordinates and color parameters. Paths are added in stages at areas of high loss following~\cite{xu2022live}. \textbf{(c)} However, the diffusion sample often fails to express all the attributes of the caption, or loses detail when vectorized. \ours{} finetunes the SVG with a latent score distillation sampling loss to improve quality and coherence.}
  \label{fig:overview}
\end{minipage} \hfill%
\begin{minipage}{.28\linewidth}
  \centering
  \includegraphics[width=\linewidth,trim={10cm 0 10cm 0},clip]{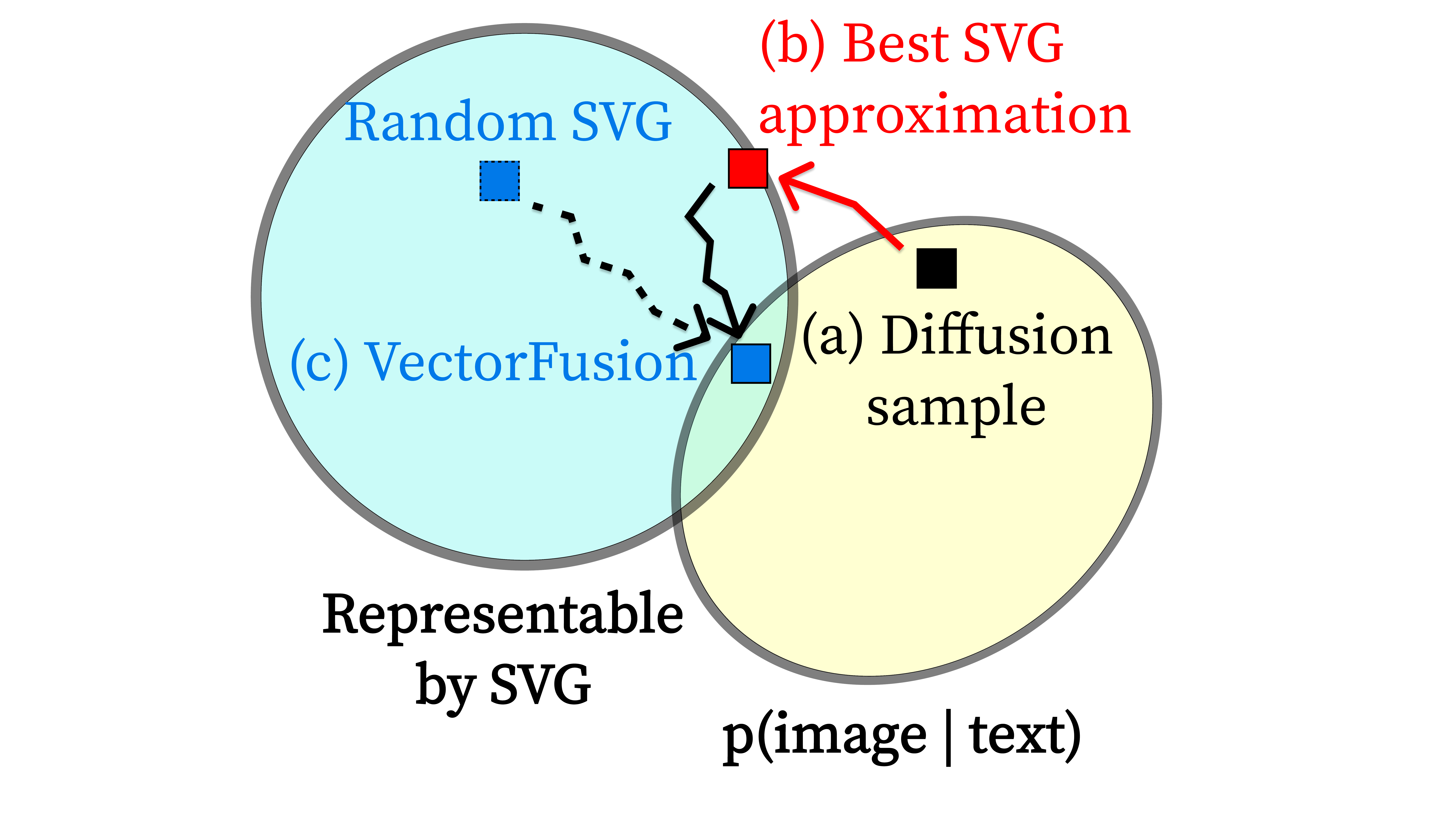}
  \captionof{figure}{Conceptual diagram motivating our approach. While {\color{red}vectorizing a rasterized diffusion sample is lossy}, \ours{} can either {\color{blue}finetune the best approximation or optimize a random SVG from scratch} to sample an SVG that is consistent with the caption.}
  \label{fig:schematic}
\end{minipage}
\end{figure*}

High resolution image synthesis is expensive. Latent diffusion models~\cite{rombach2021highresolution} train on a reduced spatial resolution by compressing $512\times 512$ images into a relatively compact $64\times 64$, 4-channel latent space with a VQGAN-like autoencoder $(E, D)$~\cite{esser2020taming}. The diffusion model $\epsilon_\phi$ is trained to model the latent space, and the decoder $D$ maps back to a high resolution raster image. We use Stable Diffusion, a popular open-source text-to-image model based on latent diffusion.

\subsection{Score distillation sampling}
\label{sec:sds}

Diffusion models can be trained on arbitrary signals, but it is easier to train them in a space where data is abundant. Standard diffusion samplers like~\eqref{eq:ddim} 
operate in the same space that the diffusion model was trained. While samplers can be modified to solve many image-to-image tasks in zero-shot such as colorization and inpainting~\cite{pmlr-v37-sohl-dickstein15, scoresde}, until recently, pretrained image diffusion models could only generate rasterized images.

In contrast, image encoders like VGG16 trained on ImageNet and CLIP (Contrastive Language–Image Pre-training)~\cite{clip} have been transferred to many modalities like mesh texture generation~\cite{mordvintsev2018differentiable}, 3D neural fields~\cite{Jain_2021_ICCV, jain2021dreamfields}, and vector graphics~\cite{jain21vector, frans21clipdraw}. Even though encoders are not generative, they can generate data with test time optimization: a loss function in the encoder's feature space is backpropagated to a learned image or function outputting images.

DreamFusion~\cite{poole2022dreamfusion} proposed an approach to use a pretrained pixel-space text-to-image diffusion model as a loss function. Their proposed Score Distillation Sampling (SDS) loss provides a way to assess the similarity between an image and a caption:
\begin{align}
    \mathcal{L}_\text{SDS} &= \mathbb{E}_{t, \epsilon} \left[ \sigma_t / \alpha_t w(t) \text{KL}(q(\x_t|g(\theta); y, t) \| p_\phi(\x_t ; y,t ))\right]. \nonumber
\end{align}
$p_\phi$ is the distribution learned by the frozen, pretrained diffusion model. $q$ is a unimodal Gaussian distribution centered at a learned mean image $g(\theta)$. 
In this manner, SDS turned sampling into an optimization problem: an image or a differentiable image parameterization (DIP)~\cite{mordvintsev2018differentiable} can be optimized with $\mathcal{L}_\text{SDS}$ to bring it toward the conditional distribution of the teacher. This is inspired by probability density distillation~\cite{Oord2018ParallelWF}. Critically, SDS only needs access to a pixel-space prior $p_\phi$, parameterized with the denoising autoencoder $\hat{\epsilon}_\phi$. It does not require access to a prior over the parameter space $\theta$. DreamFusion~\cite{poole2022dreamfusion} used SDS with the Imagen pixel space diffusion model to learn the parameters of a 3D Neural Radiance Field~\cite{mildenhall2020nerf}. In practice, SDS gives access to loss gradients, not a scalar loss:
\begin{equation} \label{eq:sds}
    \nabla_\theta \mathcal{L}_\text{SDS} = \mathbb{E}_{t, \epsilon}\left[w(t)\left(\hat\epsilon_\phi(\x_t; y, t)  - \epsilon\right) {\frac{\partial \x}{\partial \theta}}\right]
\end{equation}

\section{Method: \ours{}}
\label{sec:method}

In this section, we outline two methods for generating abstract vector representations from pretrained text-to-image diffusion models, including our full \ours{} approach.

\subsection{A baseline: text-to-image-to-vector}
\label{section:texttoimtovec}

We start by developing a two stage pipeline: sampling an image from Stable Diffusion, then vectorizing it automatically. Given text, we sample a raster image from Stable Diffusion with a Runge-Kutta solver~\cite{pndm} in 50 sampling steps with guidance scale $\omega=7.5$ (the default settings in the Diffusers library~\cite{von-platen-etal-2022-diffusers}). Naively, the diffusion model generates photographic styles and details that are very difficult to express with a few constant color SVG paths. To encourage image generations with an abstract, flat vector style, we append a suffix to the text: \textit{``minimal flat 2d vector icon. lineal color. on a white background.
trending on artstation''}. This prompt was tuned qualitatively.

Because samples can be inconsistent with captions, we sample K images and select the Stable Diffusion sample that is most consistent with the caption according to CLIP ViT-B/16~\cite{clip}. CLIP reranking was originally proposed by~\cite{dalle}. We choose K=4.

Next, we automatically trace the raster sample to convert it to an SVG using the off-the-shelf Layer-wise Image Vectorization program (LIVE)~\cite{xu2022live}. LIVE produces relatively clean SVGs by initializing paths in stages, localized to poorly recontructed, high loss regions. 
To encourage paths to explain only a single feature of the image, LIVE weights an L2 reconstruction loss by distance to the nearest path,
\begin{equation}
\mathcal{L}_\text{UDF} = \frac{1}{3} \sum_{i=1}^{w \times h} d_i' \sum_{c=1}^3 (I_{i,c} - \hat{I}_{i,c})^2
\end{equation}
where $I$ is the target image, $\hat{I}$ is the rendering, $c$ indexes RGB channels in $I$, $d_i'$ is the unsigned distance between pixel $i$, and the nearest path boundary, and $w, h$ are width and height of the image. LIVE also optimizes a self-intersection regularizer $\mathcal{L}_\text{Xing}$
\begin{equation}
\mathcal{L}_\text{Xing} =D_1 (\text{ReLU} (-D_2)) + (1-D_1) (\text{ReLU} (D_2)), \label{eq:lxing}
\end{equation}
where $D_1$ is the characteristic of the angle between two segments of a cubic Bézier path, and $D_2$ is the value of $\sin(\alpha)$ of that angle. For further clarifications of notation, please refer to LIVE~\cite{xu2022live}.

This results in a set of paths $\theta_\text{LIVE} = \{p_1, p_2, \ldots p_k\}$. Figure \ref{fig:overview}(b) shows the process of optimizing vector parameters in stages that add 8-16 paths at a time. Figure~\ref{fig:sd_comparison} shows more automatic conversions. While simple, this pipeline often creates images unsuitable for vectorization.

\subsection{Sampling vector graphics by optimization}

The pipeline in~\ref{section:texttoimtovec} is flawed since samples may not be easily representable by a set of paths. Figure~\ref{fig:schematic} illustrates the problem. Conditioned on text, a diffusion model produces samples from the distribution $p_\phi(\x | y)$. Vectorization with LIVE finds a SVG with a close L2 approximation to that image without using the caption $y$. This can lose information, and the resulting SVG graphic may no longer be coherent with the caption.

\begin{figure*}[t]
  \centering
  \includegraphics[width=0.75\linewidth]{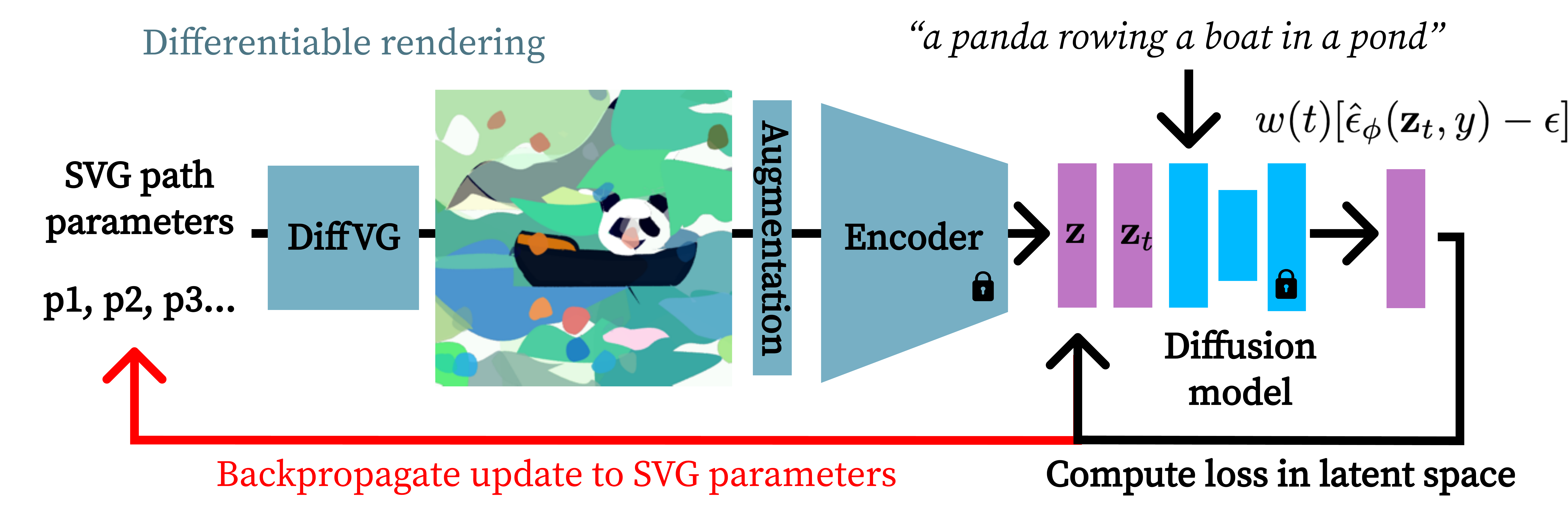}
  \captionof{figure}{An overview of \ours{}'s latent score distillation optimization procedure. We adapt Score Distillation Sampling~\cite{poole2022dreamfusion} to support a vector graphics renderer and a latent-space diffusion prior for raster images. First, we rasterize the SVG given path parameters. We apply data augmentations, encode into a latent space, compute the Score Distillation loss on the latents, and backpropagate through the encoding, augmentation and renderering procedure to update paths.}
  \label{fig:latent_score_distillation}
\end{figure*}

For \ours{}, we adapt Score Distillation Sampling to support latent diffusion models (LDM) like the open source Stable Diffusion. We initialize an SVG with a set of paths $\theta = \{p_1, p_2, \ldots p_k\}$. Every iteration, DiffVG renders a $600\times 600$ image $\x$. Like CLIPDraw~\cite{frans21clipdraw}, we augment with perspective transform and random crop to get a $512\times 512$ image $\x_\text{aug}$. Then, we propose to compute the SDS loss in latent space using the LDM encoder $E_\phi$, predicting $\z = E_\phi(\x_\text{aug})$.
For each iteration of optimization, we diffuse the latents with random noise $\z_t = \alpha_t \z + \sigma_t \epsilon$, denoise with the teacher model $\hat{\epsilon}_\phi(\mathbf{z}_t, y)$, and optimize the SDS loss using a latent-space modification of Equation~\ref{eq:sds}:
\begin{multline}
    \nabla_\theta \mathcal{L}_\text{LSDS} = \\\quad\mathbb{E}_{t, \epsilon} \left[ w(t) \Big(\hat{\epsilon}_\phi(\alpha_t \z_t + \sigma_t \epsilon, y)  - \epsilon\Big) \frac{\partial \z}{\partial \x_\text{aug}} \frac{\partial \x_\text{aug}}{\partial \theta} \right]\\
\end{multline}
Since Stable Diffusion is a discrete time model with $T=1000$ timesteps, we sample $t\sim \mathcal{U}(50, 950)$
For efficiency, we run the diffusion model $\hat{\epsilon}_\theta$ in half-precision. We found it important to compute the Jacobian of the encoder $\partial \z / \partial \x_\text{aug}$ in full FP32 precision for numerical stability. The term $\partial \x_\text{aug} / \partial \theta$ is computed with autodifferentiation through the augmentations and differentiable vector graphics rasterizer, DiffVG. $\mathcal{L}_\text{LSDS}$ can be seen as an adaptation of $\mathcal{L}_\text{SDS}$ where the rasterizer, data augmentation and frozen LDM encoder are treated as a single image generator with optimizable parameters $\theta$ for the paths. During optimization, we also regularize self-intersections with \eqref{eq:lxing}.

\subsection{Reinitializing paths}
In our most flexible setting, synthesizing flat iconographic vectors, we allow path control points, fill colors and SVG background color to be optimized. During the course of optimization, many paths learn low opacity or shrink to a small area and are unused. To encourage usage of paths and therefore more diverse and detailed images, we periodically reinitialize paths with fill-color opacity or area below a threshold. Reinitialized paths are removed from optimization and the SVG, and recreated as a randomly located and colored circle on top of existing paths. This is a hyperparameter choice, and we detail ablations and our hyperparmeters in the supplement.

\subsection{Stylizing by constraining vector representation}

Users can control the style of art generated by \ours{} by modifying the input text, or by constraining the set of primitives and parameters that can be optimized.
The choice of SVG vector primitives determines the level of abstraction of the result. We explore three settings: iconographic vector art with flat shapes, pixel art, and sketch-based line drawings.

\begin{figure}[t]
    \centering
    \includegraphics[width=\linewidth,trim={18mm 0 0 25mm},clip]{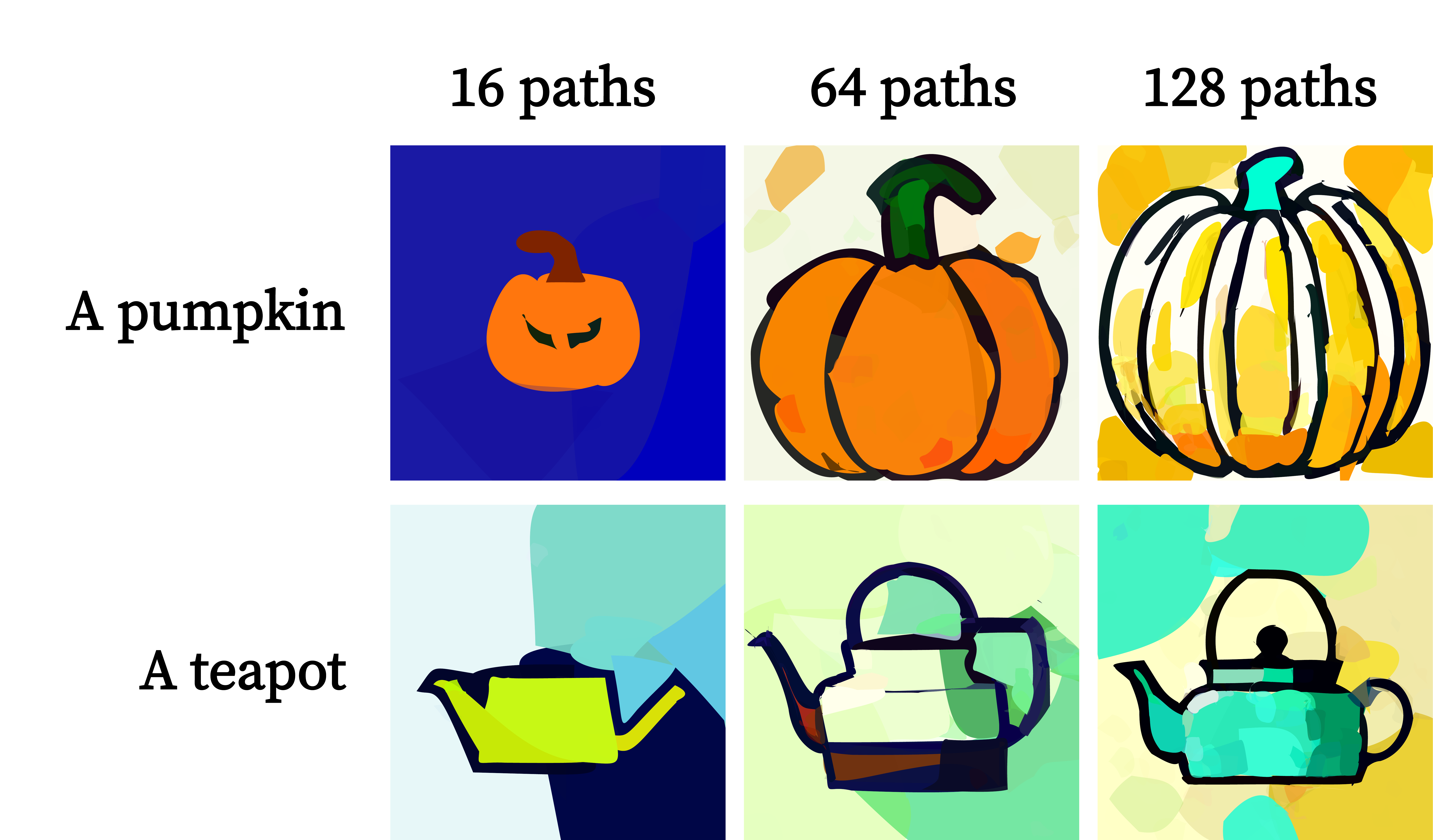}
    \caption{The number of B\'ezier paths controls the level of detail in generated vector graphics.}
    \label{fig:compare_npath}
\end{figure}

\textbf{Iconography} ~~We use closed B\'ezier paths with trainable control points and fill colors. Our final vectors have 64 paths, each with 4 segments. For \ours{} from scratch, we initialize 64 paths randomly and simultaneously, while for SD + LIVE + SDS, we initialize them iteratively during the LIVE autovectorization phase. We include details about initialization parameters in the supplement. Figure~\ref{fig:compare_npath} qualitatively compares generations using 16, 64 and 128 paths (SD + LIVE initialization with K=20 rejection samples and SDS finetuning). Using fewer paths leads to simpler, flatter icons, whereas details and more complex highlights appear with greater numbers of paths.

\textbf{Pixel art} ~~Pixel art is a popular video-game inspired style, frequently used for character and background art. While an image sample can be converted to pixel art by downsampling, this results in blurry, bland, and unrecognizable images. Thus, pixel art tries to maximize use of the available shapes to clearly convey a concept. Pixray~\cite{white21pixray} uses square SVG polygons to represent pixels and uses a CLIP-based loss following~\cite{jain21vector, frans21clipdraw}. \ours{} able to generate meaningful and aesthetic pixel art from scratch and with a Stable Diffusion initialization, shown in Fig.~\ref{fig:teaser} and Fig.~\ref{fig:pixel}. In addition to the SDS loss, we additionally penalize an L2 loss on the image scaled between -1 and 1 to combat oversaturation, detailed in the supplement. We use $32\times 32$ pixel grids.

\textbf{Sketches} ~~Line drawings are perhaps the most abstract representation of visual concepts. Line drawings such as Pablo Picasso's animal sketches are immediately recognizable, but bear little to no pixel-wise similarity to real subjects. Thus, it has been a long-standing question whether learning systems can generate semantic sketch abstractions, or if they are fixated on low-level textures. Past work includes directly training a model to output strokes like Sketch-RNN~\cite{sketchrnn}, or optimizing sketches to match a reference image in CLIP feature space~\cite{vinker2022clipasso}. As a highly constrained representation, we optimize only the control point coordinates of a set of fixed width, solid black B\'ezier curves. We use 16 strokes, each 6 pixels wide with 5 segments, randomly initialized and trained from scratch, since the diffusion model inconsistently generates minimal sketches.  More details on the training hyperparameters are included in the supplement.

\section{Experiments}
\label{sec:experiments}

In this section, we quantitatively and qualitatively evaluate the text-to-SVG synthesis capabilities of \ours{} guided by the following questions. In Section~\ref{sec:main_experiments}, we ask \textit{(1) Are SVGs generated by \ours{} consistent with representative input captions?} and \textit{(2) Does our diffusion optimization-based approach help compared to simpler baselines?} In Section~\ref{sec:experiments:clip_comparison}, we qualitatively compare \ours{}'s diffusion-based results with past CLIP-based methods. Section~\ref{sec:experiments:pixel_art} and \ref{sec:experiments:sketches} describe pixel and sketch art generations.
Overall, \ours{} performs competitively on quantitative caption consistency metrics, and qualitatively produces the most coherent and visually pleasing vectors.

\subsection{Experimental setup}

\setlength{\tabcolsep}{3pt}
\begin{figure}[t]
  \centering
  \captionof{table}{Evaluating the consistency of text-to-SVG generations using 64 primitives with input captions. Consistency is measured with CLIP R-Precision and CLIP similarity score ($\times$100). Higher is better. We compare a CLIP-based approach, CLIPDraw, with diffusion baselines: the best of K raster samples from Stable Diffusion (SD), converting the best of K samples to vectors with LIVE~\cite{xu2022live}, and \ours{} from scratch or initialized with the LIVE converted SVG. \ours{} generations are significantly more consistent with captions than Stable Diffusion samples and their automatic vector conversions. CLIPDraw is trained to maximize CLIP score, so it has artificially high scores.}
  \label{tab:main_results}
  \begin{tabular}{@{}lccccc@{}}
    \toprule
    & & \multicolumn{4}{c}{Caption consistency} \\
    & & \multicolumn{2}{c}{CLIP L/14} & \multicolumn{2}{c}{OpenCLIP H/14} \\ 
    Method & K & R-Prec & Sim & R-Prec & Sim \\ 
    \midrule
    CLIPDraw (scratch) & -- & {\bf 85.2} & 27.2 & 77.3 & {\bf 31.7} \\\midrule
    Stable Diff (\textbf{raster}) & 1 & 67.2 & 23.0 & 69.5 & 26.7 \\
    ~~+ rejection sampling & 4 & 81.3 & 24.1 & 80.5 & 28.2 \\ \midrule
    SD init + LIVE & 1 & 57.0 & 21.7 & 59.4 & 25.8 \\ 
    ~~+ rejection sampling & 4 & 69.5 & 22.9 & 65.6 & 27.6 \\
    \ours{} (scratch) & -- & 76.6 & 24.3 & 69.5 & 28.5 \\
    ~~+ SD init + LIVE & 1 & 78.1 &\underline{29.1} & \underline{78.1} & \underline{29.3} \\
    ~~+ rejection sampling & 4 & \underline{78.9} & {\bf 29.4} & {\bf 81.3} & 24.5 \\
    \bottomrule
  \end{tabular}%
\end{figure}

\begin{figure}[t]
    \centering
    \includegraphics[width=\linewidth]{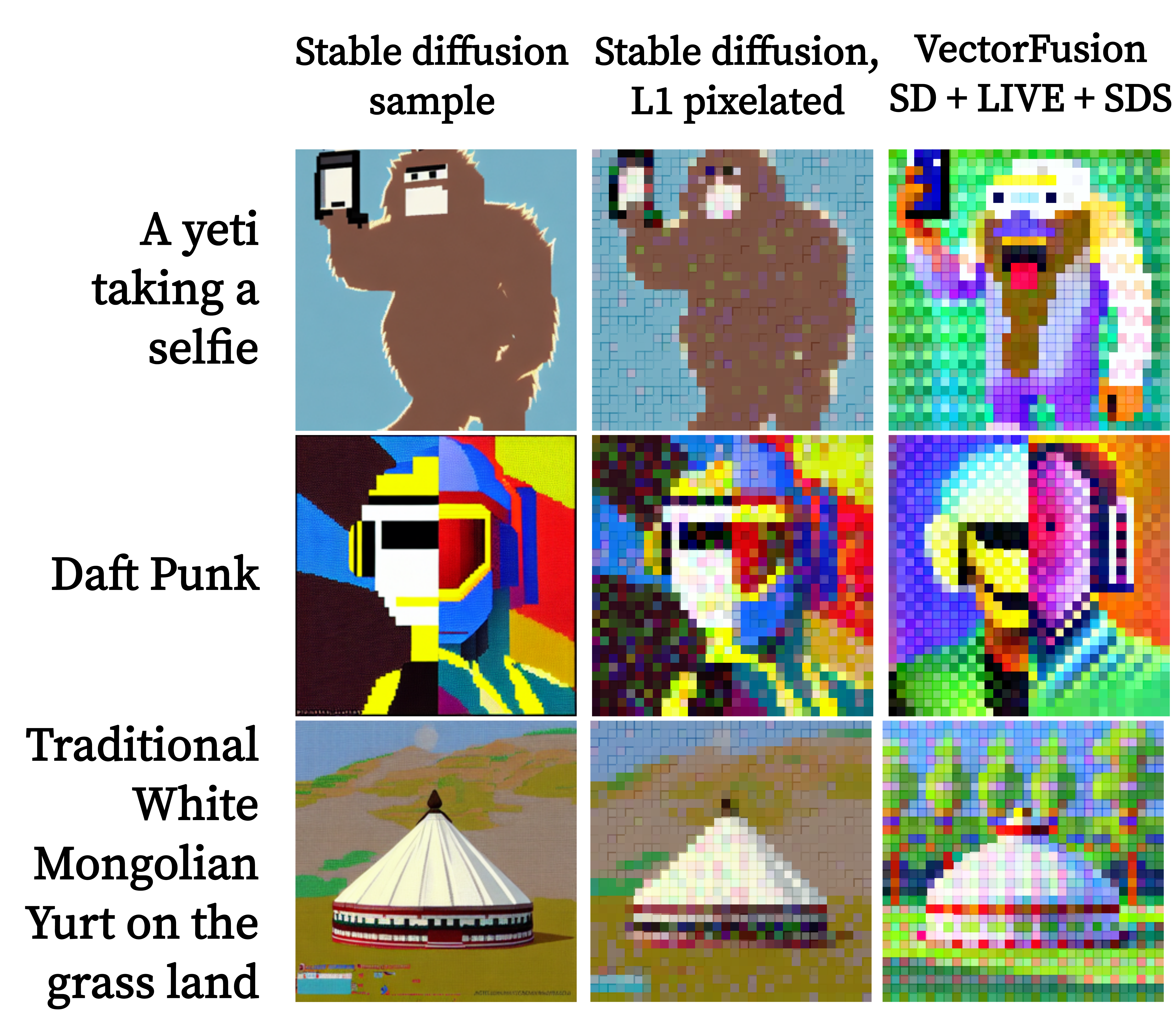}
    \caption{\textbf{\ours{} generates coherent, pleasing pixel art.} Stable Diffusion can generate a pixel art style, but has no control over the regularity and resolution of the pixel grid (left). This causes artifacts and blurring when pixelating the sample into a 32x32 grid, even with a robust L1 loss (middle). By finetuning the L1 result, \ours{} improves quality and generates an abstration that works well despite the low resolution constraint.} 
    \label{fig:pixel}
\end{figure}

It is challenging to evaluate text-to-SVG synthesis, since we do not have target, ground truth SVGs to use as a reference. We collect a diverse evaluation dataset of captions and evaluate text-SVG coherence with automated CLIP metrics. Our dataset consists of 128 captions from past work and benchmarks for text-to-SVG and text-to-image generation: prompts from CLIPDraw~\cite{frans21clipdraw} and ES-CLIP~\cite{esclip}, combined with captions from PartiPrompts~\cite{parti}, DrawBench~\cite{sr3}, DALL-E 1~\cite{dalle}, and DreamFusion~\cite{poole2022dreamfusion}. Like previous works, we calculate CLIP R-Precision and cosine similarity.

\begin{figure*}[t]
    \centering
    \includegraphics[width=\linewidth,trim={0 0 0 2.1cm},clip]{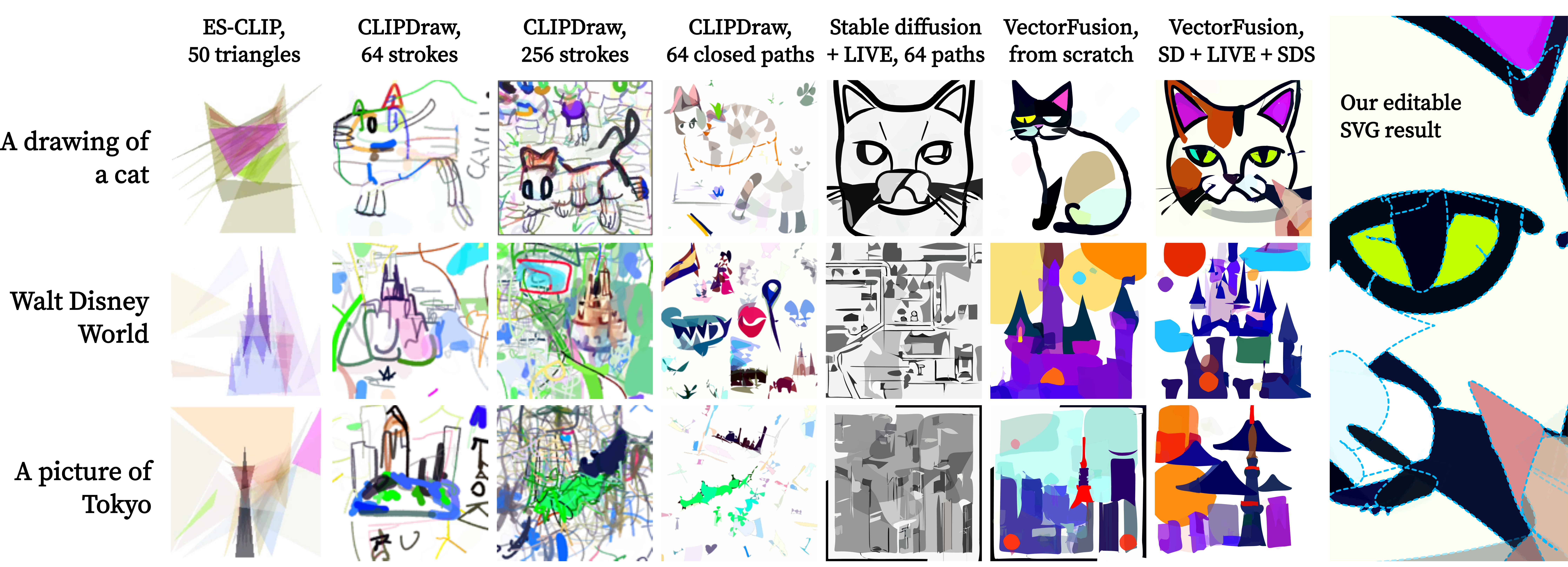}
    \caption{\ours{} produces more coherent vector art than baselines that optimize CLIP score, even with fewer paths (64 shapes). On the right, the resulting SVG can be enlarged to arbitrary scale. Individual paths are highlighted with blue dashed lines. The result can be edited intuitively by the user in design software.}
    \label{fig:clip_comparison}
\end{figure*}

\textbf{CLIP Similarity} We calculate the average cosine similarity of CLIP embeddings of generated images and the text captions used to generate them. Any prompt engineering is excluded from the reference text. As CLIP Similarity increases, pairs will generally be more consistent with each other. We note that CLIPDraw methods directly optimize CLIP similarity scores and have impressive metrics, but rendered vector graphics are sketch-like and messy unlike the more cohesive \ours{} samples. We provide examples in Figure~\ref{fig:clip_comparison}. To mitigate this effect, we also evaluate the open source Open CLIP ViT-H/14 model, which uses a different dataset for training the representations.

\textbf{CLIP R-Precision} For a more interpretable metric, we also compute CLIP Retrieval Precision. Given our dataset of captions, we calculate CLIP similarity scores for each caption and each rendered image of generated SVGs. R-Precision is the percent of SVGs with maximal CLIP Similarity with the correct input caption, among all 128.

\subsection{Evaluating caption consistency}
\label{sec:main_experiments}

As a baseline, we generate an SVG for each caption in our benchmark using CLIPDraw~\cite{frans21clipdraw} with 64 strokes and their default hyperparameters. We sample 4 raster graphics per prompt from Stable Diffusion as an oracle. These are selected amongst with CLIP reranking (rejection sampling). Stable Diffusion produces rasterized images, not SVGs, but can be evaluated as an oracle with the same metrics. We then autotrace the samples into SVGs using LIVE with 64 strokes, incrementally added in 5 stages of optimization. Finally, we generate with \ours{}, trained from scratch on 64 random paths per prompt, or initialized with LIVE.

Table~\ref{tab:main_results} shows results. Rejection sampling helps the baseline Stable Diffusion model +21.1\% on OpenCLIP R-Prec, suggesting that it is a surprisingly weak prior. LIVE SVG conversion hurts caption consistency (-14.9\% OpenCLIP R-Prec) even with 20 rejection samples compared to the raster oracle, indicating that SD images are difficult to abstract post-hoc.
In contrast, even without rejection sampling or initializing from a sample, \ours{} trained simultaneously on all random paths outperforms the best SD + LIVE baseline by +3.9\% R-Prec.
When introducing SD initialization and rejection sampling, \ours{} matches or exceeds the best LIVE baseline by +15.7\%. Thus, we note that rejection sampling consistently improves results, suggesting that a strong initialization is helpful to \ours{}, but our method is robust enough to tolerate a random init.. Our final method is competitive with or outperforms CLIPDraw (+4.0\% OpenCLIP R-Prec with reranking or +0.8\% even without using CLIP).

\subsection{Comparison with CLIP-based approaches}
\label{sec:experiments:clip_comparison}

Figure~\ref{fig:clip_comparison} qualitatively compares diffusion with CLIP-guided text-to-SVG synthesis. ES-CLIP~\cite{esclip} is an evolutionary search algorithm that searches for triangle abstractions that maximize CLIP score, whereas CLIPDraw uses gradient-based optimization. \ours{} produces much clearer, cleaner vector graphics than CLIP baselines, because we incorporate a generative prior for image appearance. However, a generative prior is not enough. Optimizing paths with the latent SDS loss $\mathcal{L}_\text{LSDS}$ (right two columns) further improves vibrancy and clarity compared to tracing Stable Diffusion samples with LIVE.

\subsection{Pixel art generation}
\label{sec:experiments:pixel_art}
\ours{} generates aesthetic and relevant pixel art. Figure~\ref{fig:teaser} shows that \ours{} from scratch can generate striking and coherent samples. Figure~\ref{fig:pixel} shows our improvements over L1-pixelated Stable Diffusion samples. We pixelate samples by minimizing an L1 loss with respect to square colors. While Stable Diffusion can provide meaningful reference images, \ours{} is able to add finer details and adopt a more characteristic pixel style.

\subsection{Sketches and line drawings}
\label{sec:experiments:sketches}
Figure~\ref{fig:teaser} includes line drawing samples. \ours{} produces recognizable and clear sketches from scratch without any image reference, even complex scenes with multiple objects. In addition, it is able to ignore distractor terms irrelevant to sketches, such as \textit{``watercolor''} or \textit{``Brightly colored''} and capture the semantic information of the caption.

\section{Discussion}
\label{sec:discussion}
We have presented \ours{}, a novel text-to-vector generative model. Without access to datasets of captioned SVGs, we use pretrained diffusion models to guide generation. The resulting abstract SVG representations can be intuitively used in existing design workflows. Our method shows the effectiveness of distilling generative models compared to using contrastive models like CLIP. In general, we are enthusiastic about the potential of scalable generative models trained in pixel space to transfer to new tasks, with interpretable, editable outputs. \ours{} provides a reference point for designing such systems.

\ours{} faces certain limitations. For instance, forward passes through the generative model are more computationally expensive than contrastive approaches due to its increased capacity. \ours{} is also inherently limited by Stable Diffusion in terms of dataset biases~\cite{laionbias} and quality, though we expect that as text-to-image models advance, \ours{} will likewise continue to improve. 

\section*{Acknowledgements}
\label{sec:ack}
We thank Paras Jain, Ben Poole, Aleksander Holynski and Dave Epstein for helpful discussions about this project. \ours{} relies upon several open source software libraries~\cite{pytorch, harris2020array, Li:2020:DVG, xu2022live, von-platen-etal-2022-diffusers, eriba2019kornia}. This work was supported in part by the BAIR Industrial Consortium.

{\small
\bibliographystyle{ieee_fullname}
\bibliography{main}

\begin{thebibliography}{10}\itemsep=-1pt

\bibitem{laionbias}
Abeba Birhane, Vinay~Uday Prabhu, and Emmanuel Kahembwe.
\newblock Multimodal datasets: misogyny, pornography, and malignant
  stereotypes, 2021.

\bibitem{esser2020taming}
Patrick Esser, Robin Rombach, and Björn Ommer.
\newblock Taming transformers for high-resolution image synthesis, 2020.

\bibitem{fernando2021genart}
Chrisantha Fernando, S.~M.~Ali Eslami, Jean-Baptiste Alayrac, Piotr Mirowski,
  Dylan Banarse, and Simon Osindero.
\newblock Generative art using neural visual grammars and dual encoders, 2021.

\bibitem{frans21clipdraw}
Kevin Frans, Lisa~B. Soros, and Olaf Witkowski.
\newblock Clipdraw: Exploring text-to-drawing synthesis through language-image
  encoders.
\newblock {\em CoRR}, abs/2106.14843, 2021.

\bibitem{goh2021multimodal}
Gabriel Goh, Nick~Cammarata †, Chelsea~Voss †, Shan Carter, Michael Petrov,
  Ludwig Schubert, Alec Radford, and Chris Olah.
\newblock Multimodal neurons in artificial neural networks.
\newblock {\em Distill}, 2021.
\newblock https://distill.pub/2021/multimodal-neurons.

\bibitem{ddpmpnp}
Alexandros Graikos, Nikolay Malkin, Nebojsa Jojic, and Dimitris Samaras.
\newblock Diffusion models as plug-and-play priors.
\newblock {\em arXiv:2206.09012}, 2022.

\bibitem{sketchrnn}
David Ha and Douglas Eck.
\newblock A neural representation of sketch drawings, 2017.

\bibitem{harris2020array}
Charles~R. Harris, K.~Jarrod Millman, St{\'{e}}fan~J. van~der Walt, Ralf
  Gommers, Pauli Virtanen, David Cournapeau, Eric Wieser, Julian Taylor,
  Sebastian Berg, Nathaniel~J. Smith, Robert Kern, Matti Picus, Stephan Hoyer,
  Marten~H. van Kerkwijk, Matthew Brett, Allan Haldane, Jaime~Fern{\'{a}}ndez
  del R{\'{i}}o, Mark Wiebe, Pearu Peterson, Pierre G{\'{e}}rard-Marchant,
  Kevin Sheppard, Tyler Reddy, Warren Weckesser, Hameer Abbasi, Christoph
  Gohlke, and Travis~E. Oliphant.
\newblock Array programming with {NumPy}.
\newblock {\em Nature}, 585(7825):357--362, Sept. 2020.

\bibitem{ddpm}
Jonathan Ho, Ajay Jain, and Pieter Abbeel.
\newblock Denoising diffusion probabilistic models.
\newblock {\em NeurIPS}, 2020.

\bibitem{classifierfree}
Jonathan Ho and Tim Salimans.
\newblock Classifier-free diffusion guidance.
\newblock {\em arXiv:2207.12598}, 2022.

\bibitem{jain21vector}
Ajay Jain.
\newblock Vectorascent: Generate vector graphics from a textual description,
  2021.

\bibitem{jain2021dreamfields}
Ajay Jain, Ben Mildenhall, Jonathan~T. Barron, Pieter Abbeel, and Ben Poole.
\newblock Zero-shot text-guided object generation with dream fields.
\newblock {\em CVPR}, 2022.

\bibitem{Jain_2021_ICCV}
Ajay Jain, Matthew Tancik, and Pieter Abbeel.
\newblock Putting nerf on a diet: Semantically consistent few-shot view
  synthesis.
\newblock In {\em Proceedings of the IEEE/CVF International Conference on
  Computer Vision (ICCV)}, pages 5885--5894, October 2021.

\bibitem{kingma2021on}
Diederik~P Kingma, Tim Salimans, Ben Poole, and Jonathan Ho.
\newblock Variational diffusion models.
\newblock {\em NeurIPS}, 2021.

\bibitem{Kingma2014AutoEncodingVB}
Diederik~P. Kingma and Max Welling.
\newblock Auto-encoding variational bayes.
\newblock {\em ICLR}, 2014.

\bibitem{Li:2020:DVG}
Tzu-Mao Li, Michal Luk\'{a}\v{c}, Gharbi Micha\"{e}l, and Jonathan
  Ragan-Kelley.
\newblock Differentiable vector graphics rasterization for editing and
  learning.
\newblock {\em ACM Trans. Graph. (Proc. SIGGRAPH Asia)}, 39(6):193:1--193:15,
  2020.

\bibitem{pndm}
Luping Liu, Yi Ren, Zhijie Lin, and Zhou Zhao.
\newblock Pseudo numerical methods for diffusion models on manifolds, 2022.

\bibitem{repaint}
Andreas Lugmayr, Martin Danelljan, Andres Romero, Fisher Yu, Radu Timofte, and
  Luc Van~Gool.
\newblock Repaint: Inpainting using denoising diffusion probabilistic models,
  2022.

\bibitem{xu2022live}
Xu Ma, Yuqian Zhou, Xingqian Xu, Bin Sun, Valerii Filev, Nikita Orlov, Yun Fu,
  and Humphrey Shi.
\newblock Towards layer-wise image vectorization.
\newblock In {\em Proceedings of the IEEE conference on computer vision and
  pattern recognition}, 2022.

\bibitem{metzer2022latentnerf}
Gal Metzer, Elad Richardson, Or Patashnik, Raja Giryes, and Daniel Cohen-Or.
\newblock Latent-nerf for shape-guided generation of 3d shapes and textures,
  2022.

\bibitem{mildenhall2020nerf}
Ben Mildenhall, Pratul~P. Srinivasan, Matthew Tancik, Jonathan~T. Barron, Ravi
  Ramamoorthi, and Ren Ng.
\newblock {NeRF}: Representing scenes as neural radiance fields for view
  synthesis.
\newblock {\em ECCV}, 2020.

\bibitem{mirowski2022clip}
Piotr Mirowski, Dylan Banarse, Mateusz Malinowski, Simon Osindero, and
  Chrisantha Fernando.
\newblock Clip-clop: Clip-guided collage and photomontage.
\newblock In {\em Proceedings of the Thirteenth International Conference on
  Computational Creativity}, 2022.

\bibitem{mordvintsev2018differentiable}
Alexander Mordvintsev, Nicola Pezzotti, Ludwig Schubert, and Chris Olah.
\newblock Differentiable image parameterizations.
\newblock {\em Distill}, 2018.

\bibitem{Nichol2022GLIDETP}
Alex Nichol, Prafulla Dhariwal, Aditya Ramesh, Pranav Shyam, Pamela Mishkin,
  Bob McGrew, Ilya Sutskever, and Mark Chen.
\newblock Glide: Towards photorealistic image generation and editing with
  text-guided diffusion models.
\newblock In {\em ICML}, 2022.

\bibitem{pytorch}
Adam Paszke, Sam Gross, Francisco Massa, Adam Lerer, James Bradbury, Gregory
  Chanan, Trevor Killeen, Zeming Lin, Natalia Gimelshein, Luca Antiga, Alban
  Desmaison, Andreas Kopf, Edward Yang, Zachary DeVito, Martin Raison, Alykhan
  Tejani, Sasank Chilamkurthy, Benoit Steiner, Lu Fang, Junjie Bai, and Soumith
  Chintala.
\newblock Pytorch: An imperative style, high-performance deep learning library.
\newblock In {\em Advances in Neural Information Processing Systems 32}, pages
  8024--8035. Curran Associates, Inc., 2019.

\bibitem{poole2022dreamfusion}
Ben Poole, Ajay Jain, Jonathan~T. Barron, and Ben Mildenhall.
\newblock Dreamfusion: Text-to-3d using 2d diffusion.
\newblock {\em arXiv}, 2022.

\bibitem{clip}
Alec Radford, Jong~Wook Kim, Chris Hallacy, Aditya Ramesh, Gabriel Goh,
  Sandhini Agarwal, Girish Sastry, Amanda Askell, Pamela Mishkin, Jack Clark,
  et~al.
\newblock Learning transferable visual models from natural language
  supervision.
\newblock {\em ICML}, 2021.

\bibitem{dalle2}
Aditya Ramesh, Prafulla Dhariwal, Alex Nichol, Casey Chu, and Mark Chen.
\newblock Hierarchical text-conditional image generation with clip latents,
  2022.

\bibitem{dalle}
Aditya Ramesh, Mikhail Pavlov, Gabriel Goh, Scott Gray, Chelsea Voss, Alec
  Radford, Mark Chen, and Ilya Sutskever.
\newblock Zero-shot text-to-image generation.
\newblock {\em ICML}, 2021.

\bibitem{eriba2019kornia}
E. Riba, D. Mishkin, D. Ponsa, E. Rublee, and G. Bradski.
\newblock Kornia: an open source differentiable computer vision library for
  pytorch.
\newblock In {\em Winter Conference on Applications of Computer Vision}, 2020.

\bibitem{rombach2021highresolution}
Robin Rombach, Andreas Blattmann, Dominik Lorenz, Patrick Esser, and Björn
  Ommer.
\newblock High-resolution image synthesis with latent diffusion models, 2021.

\bibitem{unet}
Olaf Ronneberger, Philipp Fischer, and Thomas Brox.
\newblock U-net: Convolutional networks for biomedical image segmentation,
  2015.

\bibitem{imagen}
Chitwan Saharia, William Chan, Saurabh Saxena, Lala Li, Jay Whang, Emily
  Denton, Seyed Kamyar~Seyed Ghasemipour, Burcu~Karagol Ayan, S.~Sara Mahdavi,
  Rapha~Gontijo Lopes, Tim Salimans, Jonathan Ho, David~J Fleet, and Mohammad
  Norouzi.
\newblock Photorealistic text-to-image diffusion models with deep language
  understanding.
\newblock {\em arXiv:2205.11487}, 2022.

\bibitem{sr3}
Chitwan Saharia, Jonathan Ho, William Chan, Tim Salimans, David~J. Fleet, and
  Mohammad Norouzi.
\newblock Image super-resolution via iterative refinement, 2021.

\bibitem{styleclipdraw}
Peter Schaldenbrand, Zhixuan Liu, and Jean Oh.
\newblock Styleclipdraw: Coupling content and style in text-to-drawing
  translation, 2022.

\bibitem{https://doi.org/10.48550/arxiv.1409.1556}
Karen Simonyan and Andrew Zisserman.
\newblock Very deep convolutional networks for large-scale image recognition,
  2014.

\bibitem{pmlr-v37-sohl-dickstein15}
Jascha Sohl-Dickstein, Eric Weiss, Niru Maheswaranathan, and Surya Ganguli.
\newblock Deep unsupervised learning using nonequilibrium thermodynamics.
\newblock {\em ICML}, 2015.

\bibitem{ddim}
Jiaming Song, Chenlin Meng, and Stefano Ermon.
\newblock Denoising diffusion implicit models.
\newblock {\em CoRR}, abs/2010.02502, 2020.

\bibitem{scoresde}
Yang Song, Jascha Sohl-Dickstein, Diederik~P Kingma, Abhishek Kumar, Stefano
  Ermon, and Ben Poole.
\newblock Score-based generative modeling through stochastic differential
  equations.
\newblock {\em ICLR}, 2021.

\bibitem{esclip}
Yingtao Tian and David Ha.
\newblock Modern evolution strategies for creativity: Fitting concrete images
  and abstract concepts, 2021.

\bibitem{Oord2018ParallelWF}
A{\"a}ron van~den Oord, Yazhe Li, Igor Babuschkin, Karen Simonyan, Oriol
  Vinyals, Koray Kavukcuoglu, George van~den Driessche, Edward Lockhart,
  Luis~C. Cobo, Florian Stimberg, Norman Casagrande, Dominik Grewe, Seb Noury,
  Sander Dieleman, Erich Elsen, Nal Kalchbrenner, Heiga Zen, Alex Graves, Helen
  King, Tom Walters, Dan Belov, and Demis Hassabis.
\newblock {Parallel WaveNet}: Fast high-fidelity speech synthesis.
\newblock {\em ICML}, 2018.

\bibitem{vinker2022clipasso}
Yael Vinker, Ehsan Pajouheshgar, Jessica~Y. Bo, Roman~Christian Bachmann,
  Amit~Haim Bermano, Daniel Cohen-Or, Amir Zamir, and Ariel Shamir.
\newblock Clipasso: Semantically-aware object sketching, 2022.

\bibitem{von-platen-etal-2022-diffusers}
Patrick von Platen, Suraj Patil, Anton Lozhkov, Pedro Cuenca, Nathan Lambert,
  Kashif Rasul, Mishig Davaadorj, and Thomas Wolf.
\newblock Diffusers: State-of-the-art diffusion models.
\newblock \url{https://github.com/huggingface/diffusers}, 2022.

\bibitem{white21pixray}
Tom White.
\newblock Pixray.

\bibitem{parti}
Jiahui Yu, Yuanzhong Xu, Jing~Yu Koh, Thang Luong, Gunjan Baid, Zirui Wang,
  Vijay Vasudevan, Alexander Ku, Yinfei Yang, Burcu~Karagol Ayan, Ben
  Hutchinson, Wei Han, Zarana Parekh, Xin Li, Han Zhang, Jason Baldridge, and
  Yonghui Wu.
\newblock Scaling autoregressive models for content-rich text-to-image
  generation.
\newblock {\em arXiv:2206.10789}, 2022.

\end{thebibliography}
}

\clearpage
\appendix

\section{Website with results, videos, and benchmark}

Our project website at \href{https://ajayj.com/vectorfusion}{https://ajayj.com/vectorfusion} includes videos of the optimization process and many more qualitative results in SVG format.
The benchmark used for evaluation consists of 128 diverse prompts sourced from prior work and is available at \href{https://ajayj.com/vectorfusion/svg_bench_prompts.txt}{https://ajayj.com/vectorfusion/svg\_bench\_prompts.txt}.

\section{Ablation: Reinitializing paths}

We reinitialize paths below an opacity threshold or area threshold periodically, every 50 iterations. The purpose of reinitializing small, faint paths is to encourage the usage of all paths. Path are not reinitialized for the final 200-500 iterations of optimization, so reinitialized paths have enough time to converge. Reinitialization is only applied during image-text loss computation stages. Note that for SD + LIVE + SDS, we only reinitialize for SDS finetuning, not LIVE image vectorization. For CLIPDraw, we only reinitialize paths for our closed B\'ezier path version, not for the original CLIPDraw version, which consists of open B\'ezier paths where it is difficult to measure area. We detail the hyperparameters for threshold, frequency, and number of iterations of reinitialization in Table~\ref{tab:reinit_hparams}.

\setlength{\tabcolsep}{3pt}
\begin{figure}[h]
\centering
\captionof{table}{Path reinitialization hyperparameters}
\label{tab:reinit_hparams}
  \resizebox{\linewidth}{!}{%
\begin{tabular}{lcccc}
\toprule
\multirow{2}{2cm}{\textbf{Method}} & \multirow{2}{1.4cm}{\centering \textbf{Opacity Thresh.}} & \multirow{2}{1.4cm}{\centering \textbf{Area Thresh.}} & \multirow{2}{0.7cm}{\centering \textbf{Freq.}} & \multirow{2}{2cm}{\centering \textbf{Iters}} \\\\ \midrule
SDS (from scratch)   & 0.05                       & 0                       & 50                          & 1.5/2K SDS steps              \\
SD + LIVE + SDS  & 0.05                       & 64 px$^2$                     & 50                          & 0.8/1K SDS steps               \\
CLIPDraw (icon.)            & 0.05                       & 64 px$^2$                      & 50                          & 1.8/2K CLIP steps \\ \bottomrule
\end{tabular}
}
\end{figure}

Table~\ref{tab:ablation_reinit} ablates the use of reinitialization. When optimizing random paths with SDS, reinitialization gives an absolute +3.0\% increase in R-Precision according to OpenCLIP H/14 evaluation. When initialized from a LIVE traced sample, reinitialization is quite helpful (+12.5\% R-Prec).

\begin{figure}[h]
  \centering
  \captionof{table}{Evaluating path reinitialization with 64 closed, colored B\'ezier curves, our iconographic setting. \ours{} reinitializes paths during optimization to maximize their usage. This improves caption consistency both when training randomly initialized paths with SDS (SDS w/ reinit), and when initializing with a LIVE traced Stable Diffusion sample (SD + LIVE + SDS w/ reinit).}
  \label{tab:ablation_reinit}
  \begin{tabular}{@{}lccccc@{}}
    \toprule
    & & \multicolumn{4}{c}{\textbf{Caption consistency}} \\
    & & \multicolumn{2}{c}{CLIP L/14} & \multicolumn{2}{c}{OpenCLIP H/14} \\ 
    \textbf{Method} & K & R-Prec & Sim & R-Prec & Sim \\ \midrule
    SDS & 0 & 75.0 & 24.0 & 75.0 & 28.8 \\
    ~~ w/ reinit & 0 & 78.1 & 24.1 & 78.1 & \textbf{29.3} \\ \midrule
    SD + LIVE + SDS & 4 & 64.8 & 22.6 & 68.8 & 26.7 \\
    ~~ w/ reinit & 4 & \textbf{78.9} & \textbf{29.4} & \textbf{81.3} & 24.5
    \\ \bottomrule
  \end{tabular}%
\end{figure}

\section{Ablation: Number of paths}
\begin{figure}[h]
    \centering
    \includegraphics[width=\linewidth]{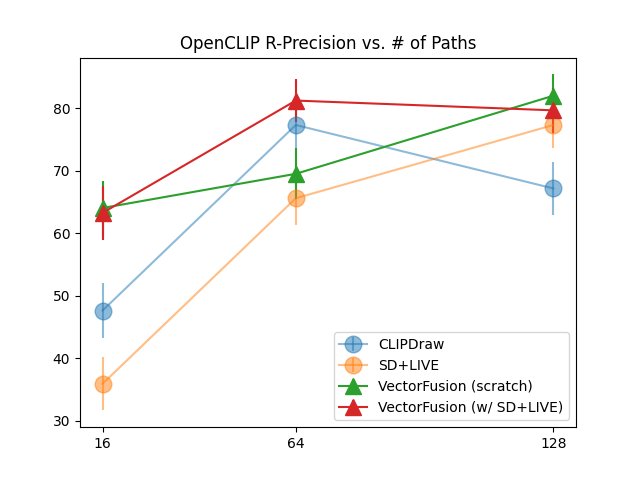}
    \caption{Increasing the number of paths generally improves our caption consistency metrics. We find 64 to be sufficient to express and optimize SVGs that are coherent with the caption.} 
    \label{fig:abl_npath}
\end{figure}

\ours{} optimizes path coordinates and colors, but the number of primitive paths is a non-differentiable hyperparameter. Vector graphics with fewer paths will be more abstract, whereas photorealism and details can be improved with many paths. In this ablation, we experiment with different number of paths. We evaluate caption consistency across path counts. For methods that use LIVE, this ablation uses a path schedule that incrementally adds 2, 4, and 10 for a total of 16 paths, and a path schedule of 8, 16, 32, and 72 for 128 total paths. We set K=4 for rejection sampling. Figure~\ref{fig:abl_npath} and Table~\ref{tab:ablation_npath} show results. Consistency improves with more paths, but there are diminishing returns.

\begin{figure}[h]
  \centering
  \captionof{table}{\textbf{Caption Consistency vs. \# Paths}. Increasing the number of paths allows for greater expressivity and caption coherency. However, it also increases memory and time complexity, and we opt for 64 paths to balance between performance and time constraints. We use rejection sampling K=4 for SD+LIVE and SD+LIVE+SDS, and we optimize open B\'ezier paths for CLIPDraw.}
  \label{tab:ablation_npath}
  \begin{tabular}{@{}lccccc@{}}
    \toprule
    & & \multicolumn{4}{c}{\textbf{Caption consistency}} \\
    & & \multicolumn{2}{c}{CLIP L/14} & \multicolumn{2}{c}{OpenCLIP H/14} \\ 
    \textbf{Method} & \# Paths & R-Prec & Sim & R-Prec & Sim \\ \midrule
    SDS (scratch) & 16 & 68.0 & 23.4 & 64.1 & 27.4 \\
    SD+LIVE & 16 & 33.6 & 19.7 & 35.9 & 22.9 \\
    SD+LIVE+SDS  & 16 & 63.3 & 22.9 & 63.3 & 27.3 \\ 
    CLIPDraw  & 16 & 58.6 & 23.9 & 47.7 & 27.4 \\
    \midrule
    SDS (scratch) & 64 & 76.6 & 24.3 & 69.5 & 28.5 \\
    SD+LIVE & 64 & 57.0 & 21.7 & 59.4 & 25.8 \\
    SD+LIVE+SDS  & 64 & 78.9 & \textbf{29.4} & 81.3 & 24.5 \\ 
    CLIPDraw  & 64 & \textbf{85.2} & 27.2 & 77.3 & \textbf{31.7} \\
    \midrule
    SDS (scratch) & 128 & 83.6 & 24.8 & \textbf{82.0} & 29.7 \\
    SD+LIVE & 128 & 77.3 & 23.7 & 77.34 & 28.7 \\
    SD+LIVE+SDS  & 128 & 78.1 & 24.8 & 79.7 & 29.7 \\ 
    CLIPDraw  & 128 & 73.4 & 25.7 & 67.2 & 30.4 \\
    \bottomrule
  \end{tabular}%
\end{figure}

\section{Ablation: Number of rejection samples}
\begin{figure}[t]
    \centering
    \includegraphics[width=\linewidth]{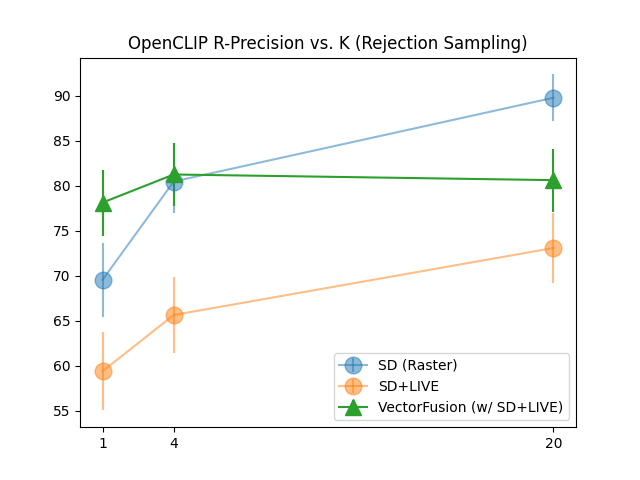}
    \caption{Coherence with the caption improves with additional rejection samples. Even with 20 rejection samples, the vectorized Stable Diffusion image baseline (SD+LIVE) still underperforms \ours{} with no rejection. \ours{} also slightly benefits from a better initialization, using 4 rejection samples of the SD initialized image.}
    \label{fig:abl_rej}
\end{figure}

In this section, we ablate on the number of Stable Diffusion samples used for rejection sampling. We include results in Figure~\ref{fig:abl_rej} and Table~\ref{tab:ablation_rej}. Rejection sampling greatly improves coherence of Stable Diffusion raster samples with the caption, since rejection explicitly maximizes a CLIP image-text similarity score. After converting the best raster sample to a vector graphic with LIVE (SD+LIVE), coherence is reduced 10-15\% in terms of OpenCLIP H/14 R-Precision. However, using more rejection samples generally improves the SD+LIVE baseline. In contrast, \ours{} is robust to the number of rejection samples. Initializing with the vectorized result after 1-4 Stable Diffusion samples is sufficient for high SVG-caption coherence.

\begin{figure}[t]
  \centering
  \captionof{table}{\textbf{Caption Consistency vs. K}. By increasing rejection sampling, we improve Stable Diffusion outputs. This improves both SD and SD+LIVE caption consistency. However, we find that \ours{} matches Stable Diffusion consistency for K=4 and retains performance for K=20. This suggests that \ours{} improves upon Stable Diffusion outputs and is robust to different initializations.}
  \label{tab:ablation_rej}
  \begin{tabular}{@{}lccccc@{}}
    \toprule
    & & \multicolumn{4}{c}{\textbf{Caption consistency}} \\
    & & \multicolumn{2}{c}{CLIP L/14} & \multicolumn{2}{c}{OpenCLIP H/14} \\ 
    \textbf{Method} & K & R-Prec & Sim & R-Prec & Sim \\ \midrule
    SD (Raster) & 1 & 67.2 & 23.0 & 69.5 & 26.7 \\
    SD+LIVE & 1 & 57.0 & 21.7 & 59.4 & 24.8 \\
    SD+LIVE+SDS & 1 & 78.1 & 24.1 & 78.1 & 29.3 \\ \midrule
    SD (Raster) & 4 & 81.3 & 24.1 & 80.5 & 28.2 \\
    SD+LIVE & 4 & 69.5 & 22.9 & 65.6 & 27.6 \\
    SD+LIVE+SDS & 4 & 78.9 & \textbf{29.4} & 81.3 & 24.5 \\
    \midrule
    SD (Raster) & 20 & \textbf{89.1} & 25.4 & \textbf{89.8} & 30.1 \\
    SD+LIVE & 20 & 71.5 & 23.6 & 73.1 & 28.5 \\
    SD+LIVE+SDS & 20 & 79.8 & 25.0 & 80.6 & \textbf{30.3} \\
    \bottomrule
  \end{tabular}%
\end{figure}

\section{Pixel Art Results}
We ablate saturation penalties and different loss objectives in Figure~\ref{tab:ablation_pix}. We use K=4 rejection samples for the initial Stable Diffusion raster image. Simply pixelating the best of K Stable Diffusion samples (SD+L1) is a straightforward way of generating pixel art, but results are often unrealistic and not as characteristic of pixel art. For example, pixelation results in blurry results since the SD sample does not use completely regular pixel grids.

Finetuning the result of pixelation with an SDS loss, and an additional L2 saturation penalty, improves OpenCLIP's R-Precision +10.2\%. Direct CLIP optimization achieves high performance on CLIP R-Precision and CLIP Similarity, but we note that like our iconographic results, CLIP optimization often yields suboptimal samples.  

\begin{figure}[t]
  \centering
  \captionof{table}{\textbf{Pixel Art.} We compare CLIP-based optimization and SDS-based optimizations. In addition, we ablate the saturation penalty, which makes pixel art more visually pleasing.}
  \label{tab:ablation_pix}
  \begin{tabular}{@{}lccccc@{}}
    \toprule
    & & \multicolumn{4}{c}{\textbf{Caption consistency}} \\
    & Sat & \multicolumn{2}{c}{CLIP L/14} & \multicolumn{2}{c}{OpenCLIP H/14} \\ 
    \textbf{Method} & Penalty& R-Prec & Sim & R-Prec & Sim \\ \midrule
    SDS (scratch) & 0 & 57.9 & 21.3 & 43.8 & 23.1  \\
    SDS (scratch) & 0.05 & 53.9 & 21.6 & 42.2 & 22.7  \\
    SD+L1 & - & 60.9 & 22.8 & 52.3 & 24.5 \\
    SD+L1+SDS & 0 & 61.8 & 23.0 & 51.6 & 24.6  \\
    SD+L1+SDS & 0.05 & 61.7 & 21.8 & 62.5 & 24.2 \\
    CLIP & - & 80.5 & 26.6 & 73.4 & 27.5 \\
    \bottomrule
  \end{tabular}%
\end{figure}

\section{Experimental hyperparameters}

In this section, we document experimental settings to foster reproducibility. In general, we find that \ours{} is robust to the choice of hyperparameters. Different settings can be used to control generation style.

\subsection{Path initialization}
\textbf{Iconographic Art} We initialize our closed B\'ezier paths with radius 20, random fill color, and opacity uniformly sampled between 0.7 and 1. Paths have 4 segments.

\textbf{Pixel Art} Pixel art is represented with a 32$\times$32 grid of square polygons. The coordinates of square vertices are not optimized. We initialize each square in the grid with a random RGB fill color and an opacity uniformly sampled between 0.7 and 1.

\textbf{Sketches} Paths are open B\'ezier curves with 5 segments each. In contrast to iconography and pixel art, which have borderless paths, sketch paths have a fixed stroke width of 6 pixels and a fixed black color. Only control point coordinates can be optimized.

\subsection{Data Augmentation}
We do not use data augmentations for SD + LIVE + SDS. We only use data augmentations for SDS trained from scratch, and CLIP baselines following~\cite{frans21clipdraw}. For SDS trained from scratch, we apply a perspective and crop augmentation. Our rasterizer renders images at a 600x600 resolution, and with 0.7 probability, we apply a perspective transform with distortion scale 0.5. Then, we apply a random 512x512 crop. All other hyperparameters are default for Kornia~\cite{eriba2019kornia}.

\subsection{Optimization}
We optimize with a batch size of 1, allowing \ours{} to run on a single low-end GPU with at least 10 GB of memory. \ours{} uses the Adam optimizer with $\beta_1=0.9$, $\beta_2=0.9$, $\epsilon = 10^{-6}$. On an NVIDIA RTX 2080ti GPU, \ours{} (SD + LIVE + SDS) takes 10-20 minutes per SVG.

For sketches and iconography, the learning rate is linearly warmed from 0.02 to 0.2 over 500 steps, then decayed with a cosine schedule to 0.05 at the end of optimization for control point coordinates. Fill colors use a 20$\times$ lower learning rate than control points, and the solid background color has a 200$\times$ lower learning rate. A higher learning rate for coordiantes can allow more structural changes.

For pixel art, we use a lower learning rate, warming from 0.00001 to 0.0001 over 1000 iterations.
We also add a weighted L2 saturation penalty on the image scaled between [-1, 1], $1/3 * \text{mean} (I_r^2 + I_b^2 + I_g^2)$, with a loss weight of 0.05. Both the lower learning rate and the L2 penalty reduced oversaturation artifacts.

\end{document}